\documentclass{IEEEtran}
\usepackage{cite}
\usepackage{lscape}
\usepackage{textcomp}
\usepackage{multirow}
\usepackage[graphicx]{realboxes}
\usepackage{adjustbox}
\usepackage{tabularx}
\usepackage[normalem]{ulem}
\usepackage{adjustbox}
\usepackage{tablefootnote}
\usepackage{xcolor}
\usepackage{verbatim}
\usepackage{algorithm}
\usepackage{algorithmicx}
\usepackage{algpseudocode}
\usepackage{mathtools}
\usepackage{amsmath,amssymb,amsfonts}
\usepackage{lscape}

\usepackage{hyperref}
\def\BibTeX{{\rm B\kern-.05em{\sc i\kern-.025em b}\kern-.08em
    T\kern-.1667em\lower.7ex\hbox{E}\kern-.125emX}}
\newcommand{\RomanNumeralCaps}[1]
    {\MakeUppercase{\romannumeral #1}}

\begin{document}

\renewcommand{\algorithmicrequire}{\textbf{Input:}}
\renewcommand{\algorithmicensure}{\textbf{Output:}}
\title{Gradual Relation Network: Decoding Intuitive Upper Extremity Movement Imaginations Based on Few-Shot EEG Learning}
\author{Kyung-Hwan~Shim, Ji-Hoon~Jeong, and~Seong-Whan~Lee*,~\IEEEmembership{Fellow,~IEEE}
\thanks{© 20xx IEEE. Personal use of this material is permitted. Permission
from IEEE must be obtained for all other uses, in any current or future media, including reprinting/republishing this material for advertising or promotional purposes, creating new collective works, for resale or redistribution to servers or lists, or reuse of any copyrighted component of this work in other works.}
\thanks{K.-H. Shim and J.-H. Jeong are with the Department of Brain and Cognitive Engineering, Korea University, Anam-dong, Seongbuk-ku, Seoul 02841, Korea. E-mail: kh\_shim@korea.ac.kr, jh\_jeong@korea.ac.kr}
\thanks{S.-W. Lee is with the Department of Artificial Intelligence, Korea University, Anam-dong, Seongbuk-ku, Seoul 02841, Korea. e-mail: sw.lee@korea.ac.kr.}
\thanks{*S.-W. Lee is the corresponding author.}}

\markboth{}
{}
\maketitle

\begin{abstract}
Brain-computer interface (BCI) is a communication tool that connects users and external devices. In a real-time BCI environment, a calibration procedure is particularly necessary for each user and each session. This procedure consumes a significant amount of time that hinders the application of a BCI system in a real-world scenario. To avoid this problem, we adopt the metric-based few-shot learning approach for decoding intuitive upper-extremity movement imagination (MI) using a gradual relation network (GRN) that can gradually consider the combination of temporal and spectral groups. We acquired the MI data of the upper-arm, forearm, and hand associated with intuitive upper-extremity movement from 25 subjects. The grand average multi-class classification results under offline analysis were 42.57\%, 55.60\%, and 80.85\% in 1-, 5-, and 25-shot settings, respectively. In addition, we could demonstrate the feasibility of intuitive MI decoding using the few-shot approach in real-time robotic arm control scenarios. Five participants could achieve a success rate of 78\% in the drinking task. Hence, we demonstrated the feasibility of the online robotic arm control with shortened calibration time by focusing on human body parts but also the accommodation of various untrained intuitive MI decoding based on the proposed GRN.
 
\end{abstract}


\section{Introduction}
\label{sec:introduction}
\IEEEPARstart{B}{rain-computer} interfaces (BCIs) allow direct communication between a human brain and an external device, without the involvement of peripheral nerves or muscular movements \cite{wolpaw2002brain, abiri2018comprehensive}. Owing to its characteristics, the BCI system has been widely used as an assistive technology or a rehabilitation system for patients with severe motor disabilities such as spinal cord injury, stroke, or amyotrophic lateral sclerosis \cite{ang2017eeg}. However, recent BCI advances have also extended their purpose to healthy people, to maximize one's physical capabilities \cite{thirdarm} or provide neural entertainment\cite{drone, lee2018high}. To this end, many researchers have adopted electroencephalogram (EEG) signals; this method offers higher portability and safety than other methods \cite{BCIReview, jeong2020decoding, Kwon_TNNLS, MI_Suk}. To control external devices, EEG-based BCI systems allow various types of experimental paradigms such as event-related potential \cite{li2018hybrid}, movement imagination (MI) \cite{Kwon_TNNLS, MI_Suk, lotze2006motor, jafarifarmand2017new}, and steady-state visual evoked potential (SSVEP) \cite{kwak2019error, leeb2015towards}. In particular, the MI-based BCI (MI-BCI) system has attracted considerable interest as it can provide direct communication between a user and a device without any external stimulus. Due to its advantages, the MI-BCI system has been specifically employed to control neuro-prosthesis (e.g., robotic arm and exoskeleton) using intuitive commands without any artificial interaction \cite{mine_2019SMC, schwarz2017decoding, jeong2020Brain}.

Although the intuitive MI-BCI system is fascinating, it has several limitations such as low decoding performance and time-consuming data acquisition process \cite{xu2019shared, shared, reduceTrial}. In this study, we focus on maintaining a sufficient decoding performance only using few amounts of EEG data. Specifically, the MI-BCI system requires a considerable amount of time to record sufficient EEG data for robust classifiers training. Owing to these inevitable BCI environments,  the MI-BCI system has considered the brain dynamics, reflecting each individual's EEG characteristics. In addition, the subjects tend toward a state of inattention state in real-time experiments due to the long calibration times for offline experiments \cite{reduceTrial, craik2019deep, singh2019reduce}.

A variety of deep learning approaches \cite{craik2019deep, eegnet, schirrmeister2017deep, tayeb2019validating} for the high performance of the MI-BCI system has drawn attention as a major advance; however, model learning still requires a large amount of EEG data. Advances in this area have a critical issue owing to the lack of large and uniform datasets. To solve this problem, we propose a gradual relation network (GRN) using only a small amount of EEG data. We adopted a few-shot learning approach that significantly contributes to the artificial intelligence (AI) field, particularly computer vision. The few-shot learning strategy can train neural networks using a few training samples, while maintaining a sufficient level of performance  \cite{relationNet, PrototypicalNetwork, Hong_Few}.  

To evaluate our proposed GRN method, we acquired various upper extremity MI data from the upper-arm, forearm, and hand motions of each subject. In addition, we applied the GRN for robotic arm control to test intuitive MI decoding with a small amount of EEG training data. Furthermore, by selecting the representative class of each body part, we could assure the user's free will on the various MI that correspond to the multivariate circumstances. We could confirm the possibility of an online brain-controlled robotic arm drinking system using intuitive MI commands with a shortened calibration time by using the GRN.

Hence, the novelty of this study can be summarized as follows: \textit{i}) This is the first attempt at adopting a few-shot learning approach to a real-time MI-BCI system; \textit{ii}) Designing a shared robotic arm control system that ensures the user's various intuitive controls rather than specifying the MI commands. \textit{iii}) The proposed GRN model outperforms conventional methods regardless of the size of the training dataset.

\section{Related Works}
A few recent BCI studies have considered deep learning techniques to recognize user intentions from EEG signals. For example, Li et al. \cite{li2018hybrid} proposed a novel hybrid network based on a restricted Boltzmann machine (RBM) for event-related potential detection. Lawhern et al. \cite{eegnet} introduced a compact CNN model (EEGNet) for BCIs. They adopted the proposed EEGNet for three typical BCI paradigms: P300 potential, movement-related cortical potentials (MRCP), and error-related negativity. The decoding performance of EEGNet was comparable to those of the conventional  methods. Schirrmeister et al. \cite{schirrmeister2017deep} demonstrated a robust MI classification for the multi-class problem using a shallow and a deep structured model ShallowConvNet and DeepConvNet, respectively. In addition, some studies presented robust EEG decoding using advanced deep learning architecture \cite{jeong2019classification}. Jiao et al. \cite{jiao2018deep} proposed a deep convolutional neural network (CNN) for mental load classification; Zhang et al., \cite{zhang2019learning} demonstrated a cross-task mental workload assessment using recurrent 3D-CNN. Lu et al. \cite {lu2017deep} proposed a novel RBM scheme for MI classification, and Zhang et al. \cite{zhang2019novel} proposed a novel deep learning approach using a data augmentation method for MI classification with small amounts of EEG data. 

Conventional BCI studies using a deep learning approach have discussed comparatively insufficient training data as one of the critical challenges \cite{craik2019deep,kaya2018large, lee2019eeg}. Constructing sufficient EEG data for training the model is realistically impossible. Therefore, we focused on employing few-shot learning approaches for studying BCIs. We hypothesized that the successful adoption of few-shot learning approaches for studying BCI can lower the calibration time for real-time scenarios and deliver a robust decoding performance using only a few data samples. Furthermore, by using the metric-based few-shot learning approach, we expect the trained model to project proper MIs to the proper sub-part of the upper extremity.

In other research fields, few-shot approaches have demonstrated successful classification and decoding for their problems. Sung et al. \cite{relationNet} presented a general and flexible framework for few-shot learning, where a classifier must be extended to new classes not observed in the training set, given only a small number of examples of each new class. Their proposed framework learns an embedding and a deep non-linear distance metric for sample items and comparing queries. Snell et al. \cite{PrototypicalNetwork} proposed a network for the problem of few-shot classification. This network learns a metric space wherein computing distances to prototype representations of each class can be used for classification. In addition, Vinyals et al. \cite{MatchNetwork} designed matching networks (MN) using only one shot learning for image classification. They showed the highest performances as 98.1\% and 98.9\%. Koch et al. \cite{koch2015siamese} proposed a deep Siamese neural network for 1-shot image recognition. They evaluated the recognition performance using the Omniglot dataset and shows comparable performance to that of a human test.

Based on these conventional studies with advanced algorithms, in this study, we designed a GRN model for robust MI decoding using only a small amount of EEG data. Furthermore, we verified that our model could contribute to real-time BCI scenarios using the robotic arm control.

\section{Data Acquisition and Experimental Setup}
\subsection{Participants}
Twenty-five healthy subjects (15 males and 10 females, aged 20-28 years, all right-handed) were recruited for the experiment. All participants were novices in the BCI system. They were sufficiently informed about the protocols of the overall experiment before it began.. As the brain signals of an individual can change according to the physiological and psychological characteristics at different times, all the volunteers were notified to attend the experiments three times at one- or two-week intervals. Subsequently, a written consent per the Declaration of Helsinki was provided. The overall experimental protocols and environments were reviewed and approved by the Institutional Review Board at Korea University (KUIRB-2020-0013-01).

\subsection{Experimental Setup}
The subject was seated comfortably on a chair before the experiment began and informed to adjust the seat 60 ($\pm 5$) cm away from the LCD monitor (refresh rate: 60 Hz; resolution: 1,920 $\times$ 1,080) as illustrated in Fig. 1(a). An EEG cap with 60 channels (ActiCap, BrainProduct GmbH, Germany) was placed on the scalp to acquire brain signals. Sixty-EEG channels followed the 10-20 international configuration and were located as followed: Fp1-2, AF5-8, AFz, F1-8, Fz, FT7-8, FC1-6, T7-8, C1-6, Cz, TP7-8, CP1-6, CPz, P1-8, Pz, PO3-4, PO7-8, POz, O1-2, Oz and Iz. The signals were recorded using the ground and reference located at FPz and FCz. The impedance of the signals were maintained below 15 k$\Omega$ during the experiments. 

\begin{figure}[!t]
\begin{center}
\includegraphics[width=\columnwidth]{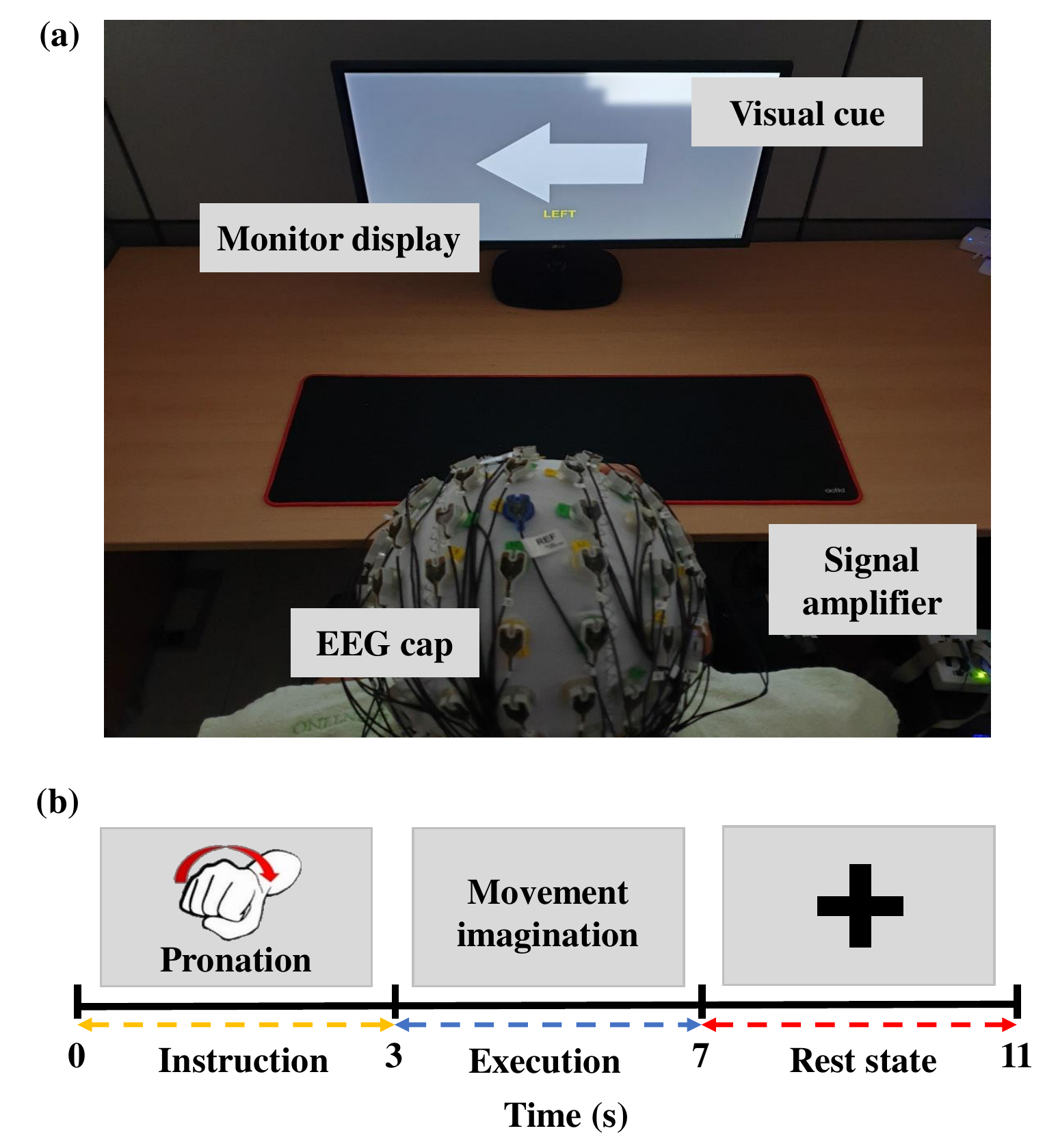}
\caption{Experimental environment. (a) Environmental configuration of the data acquisition. (b) Experiment protocols: visual instruction for 3 s, performing MI for 4 s, and resting state for 4 s with a cross sign.}
\label{fig1}
\end{center}
\end{figure}

 \subsection{Experiment Protocols}
The MI experiment protocols were designed to decode the various user intention with respect to the single upper extremity of three sub-parts the upper-arm, forearm and hand. For this reason, each session for data acquisition was divided into three parts corresponding to the sub-parts of the single upper extremity. The sequence of the three sub-parts in one session was equally distributed to collect the data on the user's condition as below.

 \subsubsection{Session 1} upper-arm $\rightarrow$ forearm $\rightarrow$ hand
 \subsubsection{Session 2} forearm $\rightarrow$ hand $\rightarrow$ upper-arm
 \subsubsection{Session 3} hand $\rightarrow$ upper-arm $\rightarrow$ forearm
 
 Each participant was provided flexible break times between the sub-parts. We collected eleven MI classes during the experiments six-classes for the upper-arm, two- for the forearm, and three- for the hand. To embrace the various intuitive MI commands, we chose one representative class for each sub-part to train the proposed GRN model. The untrained candidate classes and trials were used to verify whether the model can project into proper sub-parts. The upper-arm of the extremity typically used to select or move the object in three-dimensional space using upper-arm reaching. Therefore, six different arm-reaching MI classes were chosen as candidates among the various three-dimensional reaching. Forward upper-arm reaching MI was extracted as a representative class of upper-arm movement among six upper-arm related candidate classes. The forearm of the human body typically performed the twist; therefore, we collected two left- and right- twist MI commands as candidates and used the left-twist command as a representative class of forearm. The human hands are primarily focused on grasping the object. 
Therefore, three different grasping styles lateral, cylindrical and spherical were acquired, and the cylindrical grasp for grasping a cup was chosen as a representative class. The role of the human upper extremity cannot be limited to several specific commands. Therefore, to assure the user's free will on the robotic arm we attempted to divide the human upper extremity into three sub-parts and allow the user to perform any intuitive command. By using the metric-based approach, the GRN could successfully detect the user intention in each sub-part and allow users to perform various MIs in the online session without any constraint on the class.

The experimental paradigm consisted of three seconds of instructions, four seconds of MI, and four seconds of rest state with a fixation cross, as illustrated in Fig. 1(b). Each of the eleven classes contained fifty trials. Five hundred fifty trials for each subject at each session were acquired and 1,650 trials for each individual were obtained for the entire experiments.

\begin{algorithm*}
\caption{The overall flow of the GRN in both the training and testing phase. $D$ is all the datasets, which contain $N$ trials. $D_{s}$ is the training subset of $D_{k}$, consisting of random $n$ trials. $D_{q}$ is the untrained EEG dataset, which might contain novel candidate class or online MI data. $RandomSample(D,N)$ denotes a set of N elements chosen uniformly at random from set D, without replacement.}
\label{alg:mandelbrot}
\begin{algorithmic}[1]
\Require{Dataset $D={(x_{1},y_{1}),...,(x_{N},y_{N})}$, where each $y_{i} \in \{1,...,K\}$. $D_{k}$ denotes the subset of $D$ containing all elements $(x_{i},y_{i})$ such that $y_{i}=k$.}
\Ensure{The loss $J$ for a randomly generated training dataset on the training phase. Prediction of the unknown EEG on the other phases.}
    \If{$training$}
        \For{\texttt{$k$ in $\{1,...,K\}$}}
            \State $D_{s}=RandomSample(D_{k},n)$ \Comment{Extract $n$ trials for training for each class $k$}
            \State $c_{k} = \frac{1}{n}\sum_{(x_{j},y_{j})\in D_{s}}{f_{\theta}(x_{j})}$ \Comment{Prototypical feature of class $k$ by using embedding module $f_{\theta}$}
        \EndFor
        \State $J = 0$ \Comment{Initialize loss}
        \For{$(x_{i},y_{i})$ in $ D_{s}$}
            \For{$(x_{j},y_{j})$ in $ D_{s}$}
                \State $r_{i,j}=g_{\theta}(f_{\theta}(x_{i}), f_{\theta}(x_{j}))$ \Comment{$g_{\theta}$ represents the relation module}
                \State $J=(r_{i,j}-1(y_{i}==y_{j}))^2$ \Comment{Update MSE loss}
            \EndFor
        \EndFor
    \Else \Comment{Unknown EEG data containing the test dataset}
        \For{$x_{i}$ in $D_{q}$}
            \State $r_{k}=g_{\theta}(f_{\theta}(x_{i}), c_{k})$ \Comment{Correlation between the prototypical feature of class $k$ and unknown $x_{i}$}
        \EndFor
        \State $\sigma_{k}=\frac{exp(r_{k})}{\sum_{j}{exp(r_{j})}}$ \Comment{Softmax function to retrieve the probability}
        \State $Pred=\max_{}{\sigma_{k}}$
    \EndIf
\end{algorithmic}
\end{algorithm*}

 \section{Gradual Relation Network}
 \subsection{Data Description}
EEG signals were acquired from the 60 electrodes by sampling at 2,500 Hz, cutting off artifacts by a notch filter at 60 Hz, and band-pass filtering between 0.5 to 40 Hz, which contained most of the MI-related rhythms \cite{NNLS_CGuan}. Each individual at each session returned epoch $\times$ channel $\times$ time samples (550 $\times$ 60 $\times$ 7,500). All the signals were down-sampled to 250 Hz and 25 channels as followed: F1-4, FC1-4, C1-4, CP1-4, P1-4, Fz, FCz, Cz, CPz, and Pz were located on the sensorimotor cortex \cite{PFURTSCHELLER1997642, BCIReview, jeong2020Brain}. To extract subtle minute spatial differences we reconstructed the two-dimensional channel location from the one-dimensional location \cite{mine_2019SMC, Kwon_TNNLS}. Therefore, during training, the input to our GRN was fixed at 5$\times$5$\times$750.
 
The data acquisition process of MI consumes a relatively large amount of time to record sufficient data for robust classifier training \cite{Blankertz_NeuroImage, LotteReview, Kwon_TNNLS}. This time-consuming offline data acquisition process leads to considerable exhaustion for the user, which can deteriorate the online performance. To avoid this problem, we considered adopting few-shot learning approaches; we considered 1- and 5-shot settings as conventional few-shot approaches \cite{relationNet, MatchNetwork, Hong_Few, PrototypicalNetwork}. Furthermore, 25-shot settings (50\% of trials) were also considered to evaluate the performance of the conventional train/test division.

In few-shot settings, $n$-shot represents $n$ labeled examples for each class $k$. The training dataset $D_{s}$ labeled with class $k$ consists of $D_{s}=\{(x_{1},y_{1}),...,(x_{n},y_{n})\}$. The training strategy of the proposed GRN method follows Algorithm 1. The mean squared error loss was imported as a loss function, and all the experiments used Adam \cite{kingma2014adam} with an initial learning rate $10^{-3}$. In addition, all the models were end-to-end trained from scratch with no additional dataset.

\subsection{Overall Framework of GRN}
The GRN consisted of one encoder and one relation module to retrieve the relation score $r_{k}$ between one encoded feature and prototypical feature $c_{k}$ of class $k$, as a test phase of Algorithm 1. The encoder was trained by dataset $D_{s}$. The primary focus of this encoder was the preservation of frequency information and dimension reduction. The relation module of the GRN gathered the correlated spectral and temporal information together and compared it with the prototypical feature, $c_{k}$. By comparing the encoded feature gradually by groups, the model could accumulate the correlated frequency and temporal information more precisely. $ r_{k}=g_{\theta}(f_{\theta}(x_{i}), c_{k})$ of the test phase in Algorithm 1. depicts the overall framework of the GRN to retrieve the relation score of the input EEG signals $x_{i}$ and prototypical feature $c_{k}$.

\subsection{Encoder}
The GRN encoder contained three convolutional blocks, two temporal filters and two spatial filters as illustrated in Fig. 2. Channel-wise CNN was used to extract temporal and spectral information as reported previously \cite{NNLS_CGuan}. The receptive field of the channel-wise CNN was determined to extract the frequency information at 4 Hz and above, which had a length of (1, 1, 65). The embedding module in GRN is aimed at preserving the spectral and temporal information to be properly combined at the relation module. Herein, we developed a relation module to construct nine groups consisting of four channels; this number can be varied as a hyper-parameter. Then, we used depthwise CNN of size (5, 5, 1) to extract two frequency-specific spatial filters so that the 72 channels were extracted at the second and third layers. The spatial filter returned nine groups with eight channels that contained two spatial filters for each grouped frequency or temporal information. Lastly, the encoder of the GRN contained an additional convolutional layer to reduce the size of the embedding feature, which was a size of (1, 1, 65) but with a (1, 1, 10) stride. Each of the convolutional blocks was followed by batch normalization and exponential linear unit (ELU) non-linearization. The encoder returned 72 channels with 63 feature maps was reshaped into nine groups with eight channels ($8 \times 9 \times 63$).

While training the embedding module of GRN, the prototypical feature $c$ of each class $k$ is calculated by using $n$ labeled examples. The prototypical feature $c_{k}$ is calculated using the average encoded feature for each class $k$ as detailed in Algorithm 1.
\begin{figure}[!t]
\begin{center}
\includegraphics[width=\columnwidth]{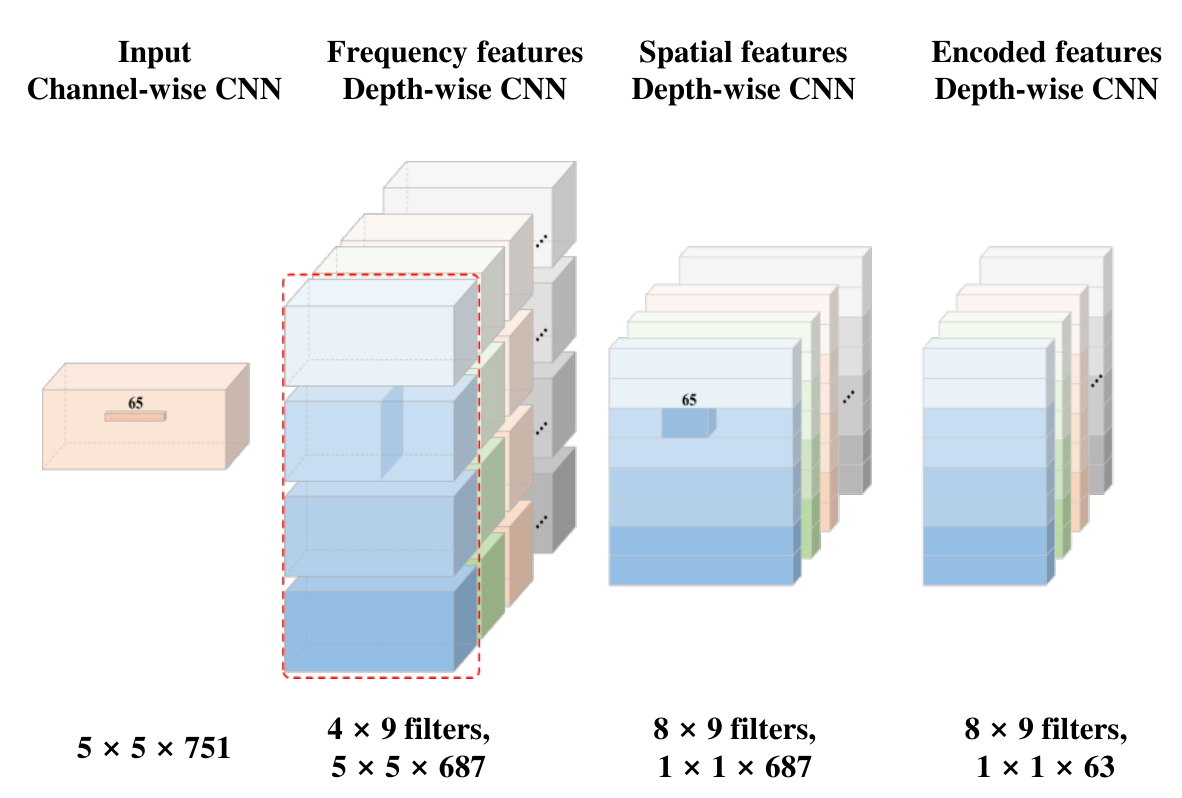}
\caption{The embedding module mainly focuses on preserving spectral and temporal information and encode the input data. The grouped data consists of the correlated feature and reallocated by the relation module.}
\label{fig3}
\end{center}
\end{figure}
\begin{figure*}[!t]
\begin{center}
\includegraphics[width=\textwidth]{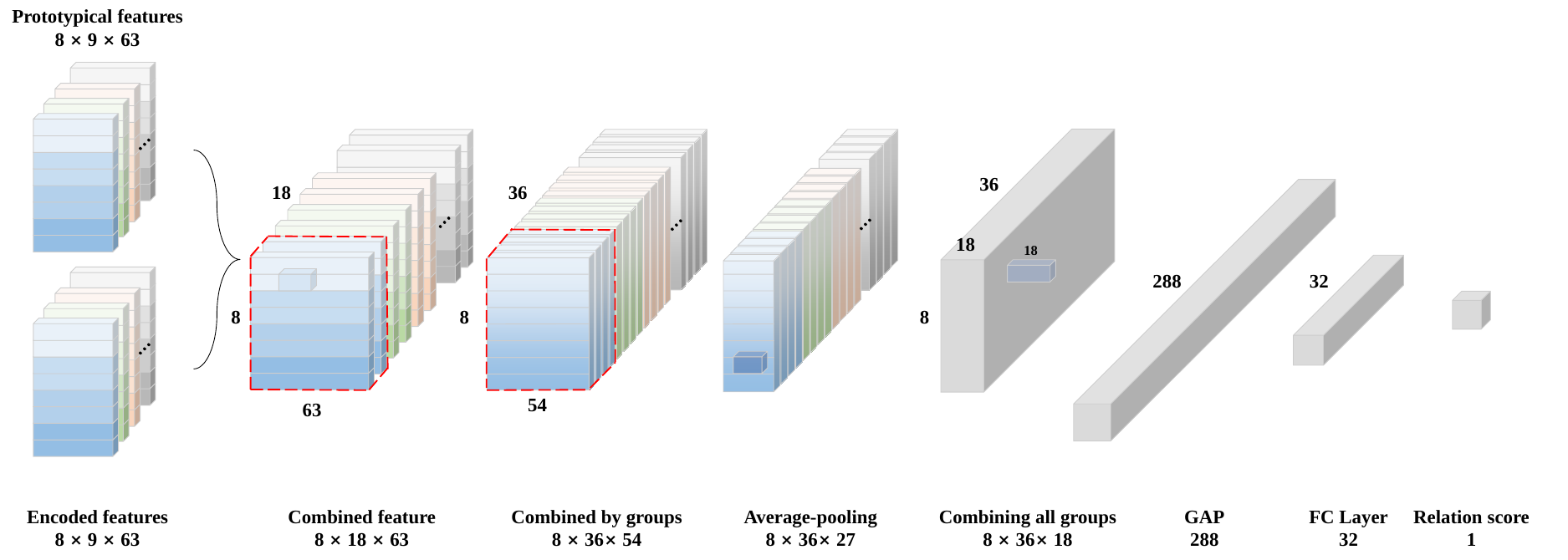}
\caption{Relation module focusing mainly on the grouping and retrieved relation score between the encoded feature and the prototypical feature of class $k$. Encoded and prototypical features are combined by group and compared gradually by the first layer of the relation module.}
\label{fig4}
\end{center}
\end{figure*}

\begin{table*}[!t]
\caption{Decoding Performance of the Proposed Method GRN on all 25 Subjects with 3 Sessions and its Average}
\renewcommand{\arraystretch}{1.3}
\centering
\resizebox{\textwidth}{!}{%
\begin{tabular}{cccccccccclccccccccc} \hline \hline
\textbf{Session}  & \multicolumn{9}{c}{Session 1}                                                          &  & \multicolumn{9}{c}{Session 2}                                                          \\ \cline{1-10} \cline{12-20}
\textbf{Training} & \multicolumn{3}{c}{1-Shot} & \multicolumn{3}{c}{5-Shot}  & \multicolumn{3}{c}{25-Shot}   &  & \multicolumn{3}{c}{1-Shot} & \multicolumn{3}{c}{5-Shot} & \multicolumn{3}{c}{25-Shot}    \\
\textbf{Subjects}     & Max.    & Avg.    & Std.   & Max.    & Avg.    & Std.    & Max.    & Avg.     & Std.   &  & Max.    & Avg.    & Std.   & Max.    & Avg.    & Std.   & Max.     & Avg.     & Std.   \\ \cline{1-10} \cline{12-20}
Sub 1       & 44.21\% & 37.95\% & 2.75 & 67.40\% & 49.62\% & 11.83 & 85.33\% & 79.60\%  & 3.72 &  & 53.74\% & 45.27\% & 4.57 & 60.74\% & 56.88\% & 3.78 & 84.00\%  & 78.53\%  & 4.51 \\
Sub 2       & 57.82\% & 42.85\% & 7.65 & 81.48\% & 69.77\% & 6.64  & 96.00\% & 93.46\%  & 2.10 &  & 44.21\% & 38.57\% & 3.05 & 54.81\% & 46.74\% & 4.52 & 90.66\%  & 85.73\%  & 3.72 \\
Sub 3       & 48.97\% & 42.10\% & 4.86 & 51.11\% & 45.25\% & 4.42  & 61.33\% & 54.93\%  & 4.68 &  & 63.94\% & 49.18\% & 9.55 & 70.37\% & 66.59\% & 2.78 & 85.33\%  & 78.80\%  & 4.40    \\
Sub 4       & 44.21\% & 37.82\% & 2.77 & 48.88\% & 42.44\% & 3.34  & 70.66\% & 63.73\%  & 5.89 &  & 40.81\% & 36.12\% & 2.07 & 47.40\% & 41.70\% & 3.31 & 69.33\%  & 61.33\%  & 5.05 \\
Sub 5       & 46.93\% & 40.00\% & 3.55 & 51.85\% & 46.22\% & 2.89  & 77.33\% & 69.73\%  & 5.43 &  & 44.21\% & 38.50\% & 2.49 & 45.18\% & 41.48\% & 1.93 & 76.00\%  & 66.4\%   & 4.94 \\
Sub 6       & 46.25\% & 38.70\% & 3.36 & 46.66\% & 41.18\% & 3.66  & 65.33\% & 60.13\%  & 3.79 &  & 48.97\% & 41.02\% & 4.73 & 68.14\% & 57.11\% & 6.54 & 86.66\%  & 75.33\%  & 7.59 \\
Sub 7       & 43.53\% & 37.07\% & 2.79 & 56.29\% & 45.33\% & 5.11  & 81.33\% & 76.26\%  & 2.65 &  & 70.74\% & 62.10\% & 8.75 & 79.25\% & 74.59\% & 3.96 & 97.33\%  & 94.93\%  & 3.08 \\
Sub 8       & 63.26\% & 58.09\% & 4.93 & 71.11\% & 63.99\% & 3.14  & 82.66\% & 74.80\%  & 5.31 &  & 45.57\% & 38.36\% & 4.06 & 60.00\% & 52.59\% & 4.98 & 85.33\%  & 77.73\%  & 4.50 \\
Sub 9       & 46.25\% & 41.22\% & 3.96 & 68.88\% & 52.44\% & 9.40  & 94.66\% & 91.46\%  & 2.32 &  & 50.34\% & 44.21\% & 3.52 & 71.85\% & 63.25\% & 5.13 & 98.66\%  & 95.33\%  & 2.54 \\
Sub 10      & 44.21\% & 38.57\% & 2.60 & 53.33\% & 46.59\% & 4.14  & 76.00\% & 72.53\%  & 3.78 &  & 48.97\% & 41.63\% & 3.96 & 67.40\% & 57.70\% & 7.20 & 92.00\%  & 87.73\%  & 3.66 \\
Sub 11      & 44.21\% & 40.00\% & 2.83 & 53.33\% & 45.70\% & 4.16  & 76.00\% & 68.53\%  & 4.47 &  & 45.57\% & 40.95\% & 3.79 & 57.03\% & 50.14\% & 4.17 & 85.33\%  & 77.73\%  & 5.09 \\
Sub 12      & 53.74\% & 44.82\% & 4.95 & 60.00\% & 56.44\% & 2.80  & 80.00\% & 74.80\%  & 4.01 &  & 61.22\% & 48.77\% & 7.76 & 79.25\% & 72.74\% & 4.09 & 94.66\%  & 88.93\%  & 4.38 \\
Sub 13      & 54.42\% & 42.38\% & 6.27 & 76.29\% & 69.70\% & 4.53  & 94.66\% & 91.60\%  & 4.04 &  & 53.74\% & 40.47\% & 4.89 & 63.70\% & 53.33\% & 5.25 & 89.33\%  & 80.93\%  & 4.50 \\
Sub 14      & 53.06\% & 39.52\% & 5.13 & 60.74\% & 51.55\% & 5.14  & 82.66\% & 74.53\%  & 4.78 &  & 49.65\% & 43.06\% & 4.91 & 63.70\% & 57.18\% & 4.61 & 93.33\%  & 87.73\%  & 3.85 \\
Sub 15      & 40.81\% & 36.80\% & 2.47 & 54.07\% & 47.48\% & 5.60  & 81.33\% & 72.53\%  & 4.05 &  & 55.10\% & 45.30\% & 4.88 & 77.03\% & 70.51\% & 7.50 & 100.00\% & 99.86\%  & 0.39 \\
Sub 16      & 40.81\% & 38.36\% & 1.29 & 50.37\% & 44.51\% & 3.60  & 58.66\% & 52.8\%   & 4.05 &  & 46.93\% & 41.76\% & 3.88 & 57.03\% & 48.44\% & 5.12 & 76.00\%  & 66.26\%  & 6.14 \\
Sub 17      & 51.02\% & 41.97\% & 4.48 & 66.66\% & 55.92\% & 7.24  & 86.66\% & 84.93\%  & 1.89 &  & 51.70\% & 40.61\% & 5.18 & 51.85\% & 47.25\% & 3.06 & 96.00\%  & 92.26\%  & 2.51 \\
Sub 18      & 50.34\% & 41.08\% & 5.25 & 62.96\% & 56.59\% & 4.74  & 92.00\% & 83.46\%  & 3.78 &  & 59.86\% & 45.64\% & 8.90 & 68.88\% & 61.33\% & 5.32 & 77.33\%  & 73.73\%  & 3.26 \\
Sub 19      & 44.89\% & 41.36\% & 2.66 & 54.07\% & 48.14\% & 4.92  & 68.00\% & 58.40\%  & 4.76 &  & 48.29\% & 41.56\% & 4.30 & 67.40\% & 57.55\% & 8.05 & 94.66\%  & 90.00\%  & 2.68 \\
Sub 20      & 41.49\% & 37.48\% & 2.62 & 54.81\% & 47.11\% & 5.94  & 93.33\% & 82.00\%  & 5.27 &  & 48.29\% & 42.99\% & 3.43 & 60.74\% & 48.22\% & 5.64 & 88.00\%  & 81.06\%  & 4.87 \\
Sub 21      & 52.17\% & 38.16\% & 2.64 & 57.03\% & 47.11\% & 6.29  & 70.66\% & 61.73\%  & 5.96 &  & 45.57\% & 40.81\% & 2.59 & 64.44\% & 52.81\% & 5.65 & 84.00\%  & 79.46\%  & 2.67 \\
Sub 22      & 52.38\% & 46.46\% & 4.47 & 60.00\% & 50.66\% & 5.29  & 93.33\% & 87.46\%  & 3.43 &  & 100.00\%& 86.19\% & 8.36 & 100.00\%& 98.51\% & 4.44 & 100.00\% & 100.00\% & 0.00 \\
Sub 23      & 48.97\% & 40.20\% & 4.17 & 60.00\% & 47.18\% & 7.43  & 76.00\% & 70.00\%  & 2.93 &  & 57.14\% & 47.95\% & 6.35 & 86.66\% & 80.14\% & 5.44 & 100.00\% & 98.66\%  & 1.57 \\
Sub 24      & 47.61\% & 40.47\% & 4.42 & 79.25\% & 61.92\% & 10.45 & 100.00\%& 100.00\% & 0.00 &  & 51.70\% & 44.96\% & 3.73 & 93.33\% & 81.85\% & 9.30 & 100.00\% & 100\%    & 0.00      \\
Sub 25      & 47.61\% & 41.29\% & 4.07 & 52.59\% & 47.77\% & 3.90  & 97.33\% & 89.33\%  & 5.29 &  & 51.70\% & 43.19\% & 5.97 & 65.92\% & 58.74\% & 5.00 & 93.33\%  & 87.60\%  & 3.44 \\\cline{1-10} \cline{12-20}
\textbf{Avg.}        & \textbf{48.37\%} & \textbf{40.99\%} & \textbf{3.88} & \textbf{59.97\%} & \textbf{51.22\%} & \textbf{5.46}  & \textbf{81.65\%} & \textbf{75.55\%}  & \textbf{3.93} &  & \textbf{53.52\%} & \textbf{45.25\%} & \textbf{5.03} & \textbf{67.28\%} & \textbf{59.89\%} & \textbf{5.07} & \textbf{89.49\%}  & \textbf{84.24\%}  & \textbf{3.58} \\ \hline \hline
            &         &         &        &         &         &         &         &          &        &  &         &         &        &         &         &        &          &          &        \\
            &         &         &        &         &         &         &         &          &        &  &         &         &        &         &         &        &          &          &        \\ \hline \hline
\textbf{Session}  & \multicolumn{9}{c}{Session 3}                                                          &  & \multicolumn{9}{c}{Avg.}                                                            \\ \cline{1-10} \cline{12-20}
\textbf{Training} & \multicolumn{3}{c}{1-Shot} & \multicolumn{3}{c}{5-Shot}  & \multicolumn{3}{c}{25-Shot}   &  & \multicolumn{3}{c}{1-Shot} & \multicolumn{3}{c}{5-Shot} & \multicolumn{3}{c}{25-Shot}    \\
\textbf{Subjects}     & Max.    & Avg.    & Std.   & Max.    & Avg.    & Std.    & Max.    & Avg.     & Std.   &  & Max.    & Avg.    & Std.   & Max.    & Avg.    & Std.   & Max.     & Avg.     & Std.   \\ \cline{1-10} \cline{12-20}
Sub 1       & 46.93\% & 41.08\% & 3.82 & 72.59\% & 61.11\% & 10.52 & 100.00\%& 98.39\%  & 1.55 &  & 48.29\% & 42.10\% & 3.71 & 66.91\% & 55.87\% & 8.71 & 89.77\%  & 85.51\%  & 3.26 \\
Sub 2       & 44.21\% & 38.77\% & 3.13 & 77.03\% & 66.88\% & 6.42  & 96.00\% & 91.46\%  & 2.93 &  & 48.75\% & 40.06\% & 4.61 & 71.11\% & 61.13\% & 5.86 & 94.22\%  & 90.22\%  & 2.92 \\
Sub 3       & 50.34\% & 41.76\% & 4.85 & 67.40\% & 58.96\% & 4.87  & 77.33\% & 69.46\%  & 4.78 &  & 54.42\% & 44.35\% & 6.42 & 62.96\% & 56.93\% & 4.02 & 74.66\%  & 67.73\%  & 4.62 \\
Sub 4       & 42.85\% & 37.55\% & 2.85 & 50.37\% & 47.18\% & 2.60  & 70.66\% & 64.53\%  & 3.27 &  & 42.63\% & 37.16\% & 2.56 & 48.88\% & 43.77\% & 3.08 & 70.00\%  & 62.53\%  & 5.47 \\
Sub 5       & 44.21\% & 37.14\% & 2.93 & 51.11\% & 43.70\% & 3.47  & 76.00\% & 71.06\%  & 2.98 &  & 45.12\% & 38.54\% & 2.99 & 49.38\% & 43.80\% & 2.76 & 76.66\%  & 68.06\%  & 5.19\\
Sub 6       & 42.85\% & 38.02\% & 2.34 & 48.88\% & 43.18\% & 3.71  & 82.66\% & 71.46\%  & 6.05 &  & 46.03\% & 39.25\% & 3.48 & 54.56\% & 47.16\% & 4.64 & 76.00\%  & 67.73\%  & 5.69 \\
Sub 7       & 46.93\% & 39.31\% & 3.78 & 49.62\% & 44.66\% & 4.12  & 81.33\% & 71.20\%  & 4.62 &  & 53.74\% & 46.16\% & 5.11 & 61.72\% & 54.86\% & 4.40 & 89.33\%  & 85.60\%  & 2.87 \\
Sub 8       & 49.65\% & 40.00\% & 4.26 & 64.44\% & 56.29\% & 5.12  & 85.33\% & 79.86\%  & 3.74 &  & 52.83\% & 45.48\% & 4.42 & 65.18\% & 57.62\% & 4.41 & 84.00\%  & 76.26\%  & 4.90 \\
Sub 9       & 59.86\% & 46.39\% & 7.38 & 92.59\% & 80.22\% & 8.54  & 100.00\%& 100.00\% & 0.00 &  & 52.15\% & 43.94\% & 4.95 & 77.77\% & 65.30\% & 7.69 & 96.66\%  & 93.40\%  & 2.43 \\
Sub 10      & 44.89\% & 38.29\% & 3.34 & 62.22\% & 54.29\% & 3.97  & 82.66\% & 77.73\%  & 3.26 &  & 46.03\% & 39.50\% & 3.30 & 60.98\% & 52.86\% & 5.10 & 84.00\%  & 80.13\%  & 3.72 \\
Sub 11      & 51.70\% & 41.76\% & 6.49 & 64.44\% & 54.66\% & 4.56  & 84.00\% & 78.26\%  & 4.62 &  & 47.16\% & 40.90\% & 4.37 & 58.27\% & 50.17\% & 4.30 & 85.33\%  & 77.73\%  & 5.09 \\
Sub 12      & 51.02\% & 43.06\% & 4.68 & 57.03\% & 51.11\% & 4.15  & 93.33\% & 86.40\%  & 4.03 &  & 55.32\% & 45.55\% & 5.80 & 65.43\% & 60.09\% & 3.68 & 94.66\%  & 88.93\%  & 4.38 \\
Sub 13      & 42.17\% & 38.29\% & 2.50 & 72.59\% & 56.96\% & 9.33  & 98.66\% & 92.26\%  & 3.99 &  & 50.11\% & 40.38\% & 4.56 & 70.86\% & 60.00\% & 6.37 & 94.00\%  & 86.60\%  & 4.24 \\
Sub 14      & 53.74\% & 45.71\% & 6.28 & 64.44\% & 54.44\% & 4.55  & 86.66\% & 80.66\%  & 3.38 &  & 52.15\% & 42.76\% & 5.44 & 62.96\% & 54.39\% & 4.77 & 87.55\%  & 80.97\%  & 4.00 \\
Sub 15      & 48.97\% & 39.59\% & 4.40 & 54.07\% & 46.81\% & 4.17  & 81.33\% & 76.40\%  & 2.60 &  & 48.29\% & 40.56\% & 3.92 & 61.72\% & 54.93\% & 5.76 & 87.55\%  & 82.93\%  & 2.35 \\
Sub 16      & 51.02\% & 44.14\% & 6.15 & 54.07\% & 50.00\% & 2.41  & 70.66\% & 62.40\%  & 4.98 &  & 46.25\% & 41.42\% & 3.77 & 53.82\% & 47.65\% & 3.71 & 64.66\%  & 57.60\%  & 4.51 \\
Sub 17      & 84.97\% & 40.40\% & 4.57 & 67.40\% & 52.96\% & 5.48  & 92.00\% & 85.46\%  & 4.31 &  & 62.56\% & 40.99\% & 4.74 & 61.97\% & 52.04\% & 5.26 & 89.33\%  & 85.20\%  & 3.10 \\
Sub 18      & 53.06\% & 43.53\% & 6.77 & 66.66\% & 60.96\% & 4.62  & 92.00\% & 86.26\%  & 3.32 &  & 54.42\% & 43.42\% & 6.97 & 66.17\% & 59.62\% & 4.90 & 92.00\%  & 84.86\%  & 3.55 \\
Sub 19      & 46.93\% & 40.88\% & 4.16 & 65.92\% & 57.40\% & 5.05  & 90.66\% & 86.00\%  & 2.93 &  & 46.71\% & 41.26\% & 3.71 & 62.46\% & 54.37\% & 6.01 & 79.33\%  & 72.20\%  & 3.84 \\
Sub 20      & 57.82\% & 44.69\% & 6.72 & 82.22\% & 65.11\% & 9.68  & 100.00\%& 98.26\%  & 1.53 &  & 49.20\% & 41.72\% & 4.26 & 65.92\% & 53.48\% & 7.09 & 90.66\%  & 81.53\%  & 5.07 \\
Sub 21      & 40.81\% & 37.82\% & 2.23 & 62.96\% & 53.70\% & 7.49  & 90.66\% & 80.80\%  & 4.54 &  & 46.19\% & 38.93\% & 2.49 & 61.48\% & 51.20\% & 6.47 & 77.33\%  & 70.60\%  & 4.32 \\
Sub 22      & 55.10\% & 45.71\% & 6.76 & 63.70\% & 59.11\% & 4.32  & 100.00\%& 98.13\%  & 1.48 &  & 69.16\% & 59.45\% & 6.53 & 74.56\% & 69.43\% & 4.68 & 97.77\%  & 95.20\%  & 1.64 \\
Sub 23      & 52.38\% & 41.02\% & 5.83 & 54.81\% & 47.55\% & 5.40  & 88.00\% & 79.33\%  & 5.78 &  & 52.83\% & 43.06\% & 5.45 & 67.16\% & 58.29\% & 6.09 & 88.00\%  & 84.33\%  & 2.25 \\
Sub 24      & 55.78\% & 44.08\% & 5.10 & 69.62\% & 60.81\% & 6.46  & 96.00\% & 90.13\%  & 3.73 &  & 51.70\% & 43.17\% & 4.42 & 80.74\% & 68.19\% & 8.74 & 100.00\% & 100.00\% & 0.00 \\
Sub 25      & 59.18\% & 48.16\% & 6.64 & 72.59\% & 64.22\% & 7.54  & 97.33\% & 93.20\%  & 3.06 &  & 52.83\% & 44.21\% & 5.56 & 63.70\% & 56.91\% & 5.48 & 95.33\%  & 88.46\%  & 4.38 \\ \cline{1-10} \cline{12-20}
\textbf{Avg.}        & \textbf{51.09\%} & \textbf{41.49\%} & \textbf{4.72} & \textbf{64.35\%} & \textbf{55.69\%} & \textbf{5.54}  & \textbf{88.53\%} & \textbf{82.76\%}  & \textbf{3.50} &  & \textbf{50.99\%} & \textbf{42.57\%} & \textbf{4.54} & \textbf{63.87\%} & \textbf{55.60\%} & \textbf{5.36} & \textbf{86.56\%}  & \textbf{80.85\%}  & \textbf{3.67} \\ \hline \hline
\end{tabular}%
}
\end{table*}

 \subsection{Relation Module}
Each of the encoders returned nine groups each containing eight channels with 63 feature maps. The relation module primarily focused on grouping the encoded features and retrieving the relation scores between the encoded features and the prototypical feature of class $k$. Both encoded and prototypical features were combined by a group and were compared gradually at the first layer of the relation module. By comparing those encoded features gradually, the relation module could compare various spectral and temporal feature groups (depicted in Fig. 7).

Specifically, one encoded feature from the unknown EEG signals and prototypical features of class $k$ were concatenated group by group alternately as illustrated in Fig. 3. The group of encoded features from the respective encoders were combined by the convolutional layer. The first convolutional layer of the relation module has the role of combining two clustered groups. Each clustered group was calculated by the convolutional layer size of (1, 10) receptive field and 32 channels were returned for each concatenated group to retrieve more information between the two groups. After this feature combination according to groups, average pooling was performed over a (1, 2) window, with stride 2. The second layer of the relation module was used to combine all the nine compared groups using a convolutional layer, with a size of (1, 10) and returned 288 filters with 18 feature maps. Global average pooling (GAP) was used to prevent the overfitting problem of the fully-connected (FC) layer, rather than vectorizing and constructing a 5,184 feature vector ($ feature \,size\, (288) \times feature\,map\, (18)$). The retrieved feature vector (with size 288) passed two FC-layer and sigmoid functions to extract a relation score between two encoded features. 

To predict the unknown EEG signals in the sub-part of the upper extremity, the relation score $r_{k}$ of each sub-part $k$ becomes an input of softmax to retrieve the probability distribution $\sigma_{k}$, and the maximum value becomes the prediction of the unknown EEG signals as a test phase of Algorithm 1.

\begin{figure*}[!t]
\begin{center}
\includegraphics[width=\textwidth]{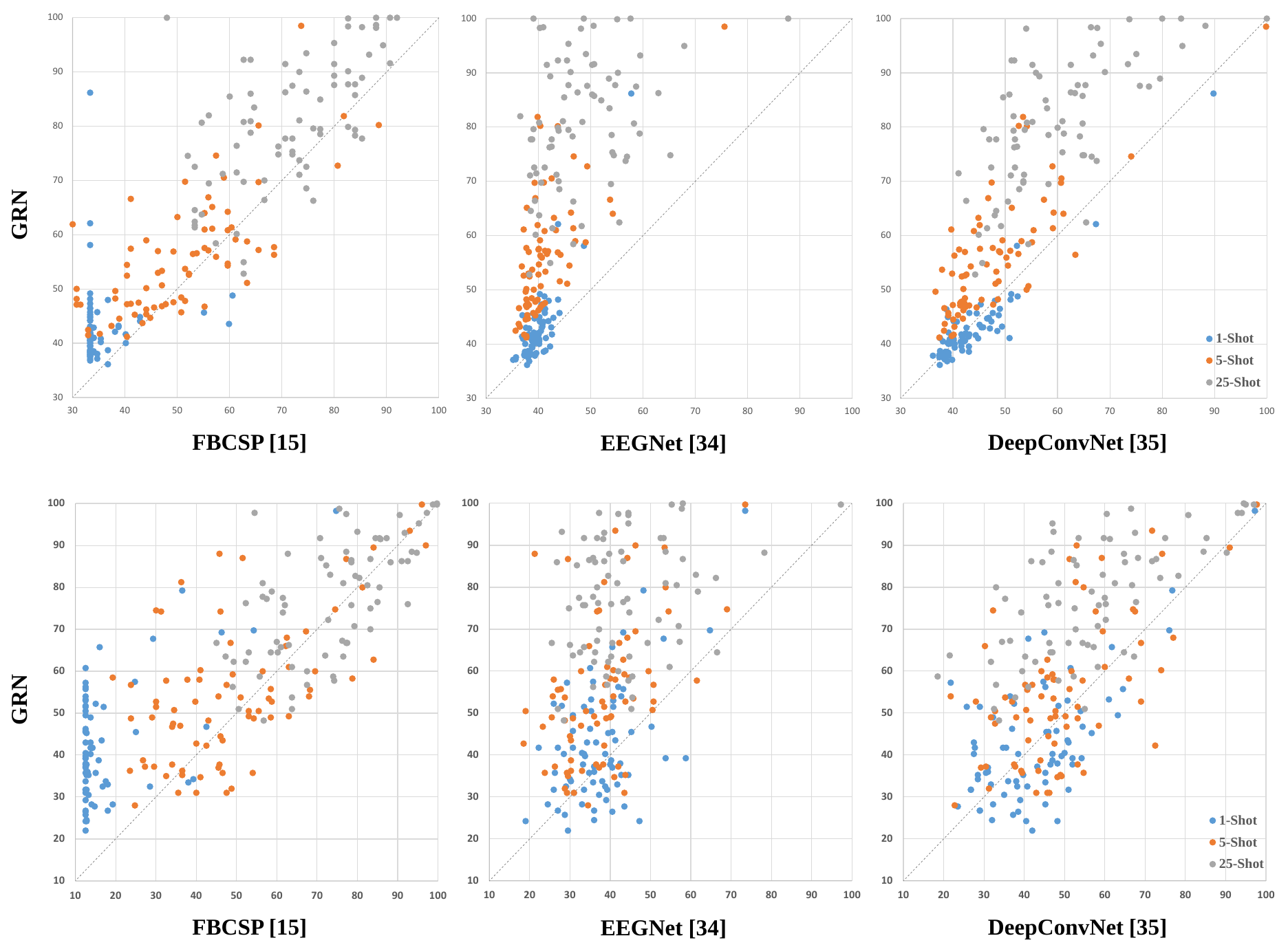}
\caption{Performance comparison of the proposed method GRN and conventional methods using 1-, 5-, and 25-shot settings. Top-row: performance comparison with FBCSP, EEGNet, and DeepConvNet on the representative classes. Bottom-row: performance comparison with FBCSP, EEGNet, and DeepConvNet on the candidate classes. The horizontal axis is the decoding accuracies of the other methods, and the vertical axis is the decoding accuracies of the GRN.}
\label{fig5}
\end{center}
\end{figure*}

\subsection{Performance Evaluation}
\subsubsection{Offline Experiment}
The performance of GRN was evaluated on two different datasets, representative classes, and candidate classes using the same model trained by the representative classes. Additionally, as offline data acquisition has the purpose of online robotic arm control in the BCI system, we also performed an online evaluation of drinking water within five training data. Therefore, we proceeded to follow three performance evaluations.

\begin{enumerate}
    \item Performance evaluation on the representative classes with 1-shot, 5-shot, and 25-shot settings.\label{item:1}
    \item Performance evaluation on the intuitive candidate classes that may occur during the online experiment.\label{item:2}
    \item System performance of robotic arm manipulation to drink water during the online experiment.\label{item:3}
\end{enumerate}

To control the robotic arm more intuitively and verify the robustness of the model, the overall dataset contained three representative classes and eight candidate classes. The eight candidate classes consisted of classes that the user may use while controlling the robotic arm.

EEG is a non-stationary signal, the statistics of which vary across time and trial \cite{review_2019, cole2019cycle, gramfort2013time}. The performance can be varied according to the composition of the training datasets. Therefore, we trained the model by using ten different combinations of datasets and retrieved the average and standard deviation as presented in Table \RomanNumeralCaps{1}. The average performance on 10 different datasets was compared with those of the conventional methods, and the comparison is presented in the first row of Fig. 4.

\subsubsection{Online Experiment}
The purpose of the time-consuming data acquisition process in the BCI system is to control or communicate with external devices. We propose the robotic arm control system with a GRN trained by a small amount of EEG data to avoid the exhaustion of the user on the offline data acquisition process. Therefore, we performed the online drinking task to prove the feasibility of the online robotic arm control system by using a small amount of data. 

Five participants (Sub 2, Sub 9, Sub 13, Sub 22, and Sub 24), who outperformed 60\% were selected as the participants in the online robotic arm control. Sub 12 was absent for the online robotic arm control session because of the personal issue. After 10 trials of the offline data acquisition process per class, five participants performed the online drinking task in the experimental environment as illustrated in Fig. 5. 

The paradigm of the online robotic arm control was developed in a similar manner to that of the offline data acquisition process but with a 5-seconds long MI signal acquisition; the instruction and rest stage was substituted by the robotic arm manipulation stage. The 5-seconds long of the MI signal acquisition stage was divided into five sliding windows, each 3-seconds long. A detailed description and the results of the online experiment are presented below for the drinking task in the Experimental Results section.

\section{Experimental Results}
\subsection{Performance Evaluation of the Representative Classes}
Table I presents the averaged decoding accuracies across all subjects by using the proposed method of the representative classes. In session 1, twenty-five subjects show 40.99\% ($\pm3.88$), 51.22\% ($\pm 5.46$), and 75.55\% ($\pm3.93$) when $n=1,5,$ and $25$, respectively. 53.52\% ($\pm5.03$), 59.89\% ($\pm5.07$), and 84.24\% ($\pm3.58$) for session 2; 41.49\% ($\pm4.72$), 55.69\% ($\pm5.54$), and 82.76\% ($\pm3.50$) for session 3. The overall performance when $n=1,5,$ and $25$ were 42.57\% ($\pm4.54$), 55.60\% ($\pm5.36$), and 80.85\% ($\pm3.67$) respectively. The standard deviation in Table \RomanNumeralCaps{1} represents the performance variation according to different training datasets of an individual. As EEG signals are non-stationary, the performance of the models trained by 1- and 5-shot settings were easily affected by the composition of the training set and exhibited comparatively large variances \cite{review_2019, cole2019cycle, gramfort2013time}.

\begin{table*}[!t]
\caption{Performance Comparison of Both Representative and Candidate Classes}
\renewcommand{\arraystretch}{1.5}
\centering
\resizebox{\textwidth}{!}{
\begin{tabular}{cc|ccccccccc} \hline \hline
\multirow{2}{*}{Dataset}               & \multirow{2}{*}{Methods} & \multicolumn{3}{c}{1-Shot}       & \multicolumn{3}{c}{5-Shot}       & \multicolumn{3}{c}{25-Shot}      \\ 
                                       &                                            & Avg.    & Model Std. & Sub. Std. & Avg.    & Model Std. & Sub. Std. & Avg.    & Model Std. & Sub. Std. \\ \hline
\multirow{4}{*}{Representative classes} & FBCSP\cite{fbcsp}                          & 35.28\% & 2.80       & 5.25      & 50.35\% & 4.48       & 12.49     & 71.58\% & 4.23       & 11.61     \\
                                       & EEGNet\cite{eegnet}                        & 39.85\% & 3.97       & 3.00      & 41.09\% & 3.93       & 5.53      & 48.04\% & 4.82       & 8.57      \\
                                       & DeepConvNet\cite{schirrmeister2017deep}    & 43.31\% & 4.45       & 7.40      & 47.60\% & 4.24       & 9.66      & 58.95\% & 4.97       & 12.96     \\
                                       & Single-grouped GRN                                 & 43.91\% & 4.46       & 6.62      & 51.38\% &5.32        & 6.10      & 61.35\% & 3.72       & 13.87     \\
                                       & \textbf{GRN}                               & \textbf{42.57\%} & \textbf{4.54} & \textbf{6.69}  & \textbf{55.60\%} & \textbf{5.36}  & \textbf{10.80}& \textbf{80.85\%} & \textbf{3.67}       & \textbf{12.03}     \\ \hline
\multirow{4}{*}{Candidate classes}       & FBCSP\cite{fbcsp}                          & 17.36\% & -          & 10.97     & 48.77\% & -          & 18.22     & 73.22\% & -          & 14.54     \\
                                       & EEGNet\cite{eegnet}                        & 37.29\% & -          & 9.09      & 37.96\% & -          & 10.4      & 43.54\% & -          & 12.34     \\
                                       & DeepConvNet\cite{schirrmeister2017deep}    & 42.91\% & -          & 13.21     & 49.11\% & -          & 14.37     & 56.59\% & -          & 17.67     \\
                                       & Single-grouped GRN                                 & 41.91\% & -          & 14.87     & 51.10\% & -          & 15.84     & 64.81\% & -          & 11.01     \\
                                       & \textbf{GRN}                               & \textbf{42.29\%} & -          & \textbf{13.77}     & \textbf{54.71\%} & -          & \textbf{16.95}     & \textbf{77.12\%} & -          & \textbf{13.99}    \\ \hline \hline
\end{tabular}%
}
\end{table*}

\begin{table}[!t]
\caption{Online Success Rate on Drinking Task Under Robotic Arm Control Environment}
\renewcommand{\arraystretch}{1.5}
\centering
\resizebox{\columnwidth}{!}{%
\begin{tabular}{ccccccc} \hline \hline
 & Sub 2 & Sub 9 & Sub 13 & Sub 22 & Sub 24 & \textbf{Avg.} \\ \hline
Success rate & 70\% & 80\% & 70\% & 80\% & 90\% & \textbf{78\%} \\
\multirow{2}{*}{Control time (s)} & 56.36  & 62.06 & 74.72 & 64.59 & 55.73 & \textbf{62.69} \\ 
                                  & ($\pm16.47$)  & ($\pm19.98$) & ($\pm15.16$) & ($\pm20.64$) &($\pm18.35$) & \textbf{($\pm18.12$)} \\ 
\multirow{2}{*}{Commands} & 8.9         & 9.8           & 11.8          & 10.2          & 8.8           & \textbf{9.9} \\ 
                                  &($\pm2.60$) &($\pm3.15$)   &($\pm2.39$)   &($\pm3.25$)   &($\pm2.89$)   &\textbf{($\pm2.86$)} \\ \hline \hline
\end{tabular}%
}
\end{table}

The first row of Fig. 4. shows the scatter plots that indicate the performance comparison between individual subjects and sessions. The x-axis represents the classification performance of the conventional methods, whereas the y-axis represents the performance of the proposed GRN. The blue, orange and grey points of the scatter plot represents the performances of 1-, 5-, and 25-shot settings, respectively. The proposed methods outperformed FBCSP by 7.29\%, 5.25\% and 9.27\% and the conventional deep-learning methods by -0.74\%, 8\%, and 21.9\% for 1-, 5-, and 25-shot settings respectively. The GRN could outperform both conventional machine learning and deep learning methods for all the training amounts except the 1-shot deep learning approaches.

Table \RomanNumeralCaps{2} compares the average performances of all the subjects at all sessions and the following two standard deviations represent the standard deviation of the model by the trainset and the standard deviation of all 25 subjects. The GRN with the single group outperformed the conventional deep learning methods on the 1-, 5-, and 25-shot settings by 0.6\%, 3.75\%, and 2.4\%, respectively. On the contrary, the GRN with the single group could not outperform FBCSP in the 25-shot setting. In could only surpass the performance on the 1- and 5-shot settings by 8.63\% and 1.03\%, respectively. The single grouped GRN exhibited a performance degradation by 10.23\% in the 25-shot setting, in comparison with FBCSP.

\subsection{Performance Evaluation of the Candidate Classes}
Collecting intuitive command in various environments is difficult, and the candidate classes were used to verify whether the model could afford and allocate the untrained intuitive classes to the proper sub-part of the upper extremity. This relatively high performances of the candidate classes allowed us to design an intuitive online robotic arm control system but without additional training.

Table \RomanNumeralCaps{2} and the second row of Fig. 4. compare the performances of the candidate datasets. The model with the highest accuracy among the ten models trained by representative classes was used to retrieve the results of the candidate classes. The proposed method outperformed FBCSP by 24.93\%, 5.94\%, and 3.9\% and the conventional deep learning approaches by -0.62\%, 5.6\%, and 20.53\% in the 1-, 5-, 25-shot settings, respectively. Even the deep learning approaches contained larger parameters and outperformed FBCSP on small amounts of training data, but it could not outperform it in the 25-shot setting. However, the proposed GRN method could outperform both deep learning and FBCSP regardless of the training amounts except in the 1-shot setting.

The GRN outperformed both representative and candidate classes on all the training settings, except in the 1-shot setting. The standard deviation of the subjects in the candidate classes was larger than that in the representative classes. However, performance degradation in the untrained candidate classes was not significant. Therefore, we could design the online experiment to perform intuitive MI rather than the representative commands acquired in the data acquisition process.

\begin{figure*}[!t]
\begin{center}
\includegraphics[width=\textwidth]{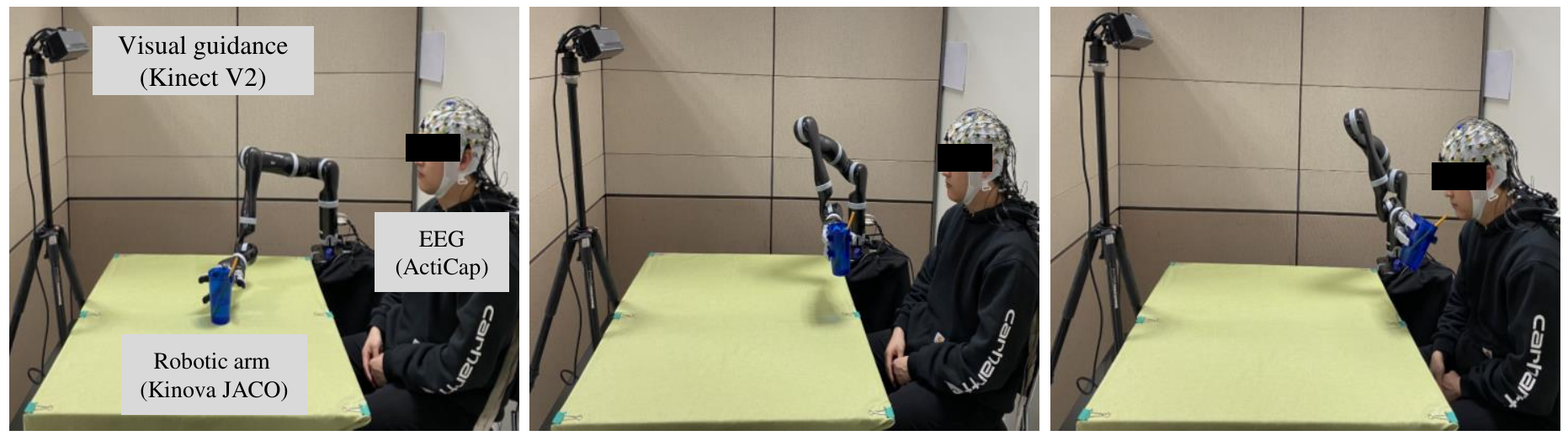}
\caption{Online experiment environments for drinking task by using a robotic arm. To perform the high-level task, the subjects imagined upper-arm reaching, hand grasping, and forearm twisting, sequentially, as the intuitive commands.}
\label{fig2}
\end{center}
\end{figure*}

\subsection{Online Robotic Arm Manipulation for Drinking Task}

The purpose of data acquisition in the BCI domain was to control or communicate with external devices; however, the long calibration time of the BCI system led to considerable distortion of the brain signals of the user, which in turn may have led to performance deterioration. Therefore, we acquired 10 trials for each representative class; five trials were allocated for training and five for the validation. This shortened calibration time allowed for avoiding the physical burden to the user. Furthermore, as we verified how candidate classes can be effectively allocated to the sub-parts in the previous section, for this reason, we instructed the user to perform the various intuitive MIs to control the robotic arm not just the class acquired by the data acquisition process.


Because of the safety issue, five subjects (Sub 2, Sub 9, Sub 13, Sub 22, and Sub  25) who exhibited higher performance than 60\% on three offline sessions attended the online robotic arm control session. The user of the system was instructed to perform ten water drinking tasks using the robotic arm, as illustrated in Fig. 5. A beep sound was provided to alert the start of online signal acquisition. The 5 seconds, starting from the beep sound, was divided into 3 seconds sliding windows with 500 millisecond strides. Each of the 3 seconds sliding windows $wind$ returned a relation score  $\sigma_{(wind,k)}$ for each class $k$ from the GRN. The highest prediction value using Equation (1) becomes a command to control the robotic arm. To assist the drinking task by using the robotic arm, each of the decoded user intentions related to the sub-part corresponds to the role of drinking. Upper-arm-related command from the user allows the robotic arm to locate adjacent to the object, the hand-related command allows the robotic arm to grasp the adjacent object and the forearm-related command of the user allows the robotic arm to tilt the robotic hand (Fig. 5).

\begin{equation}
    cmd=\max_{k}{\frac{1}{5}\sum_{wind}{\sigma_{(wind,k)}}}
\end{equation}

The user of the system was instructed to perform drink water 10 times using the robotic arm. During this online experiment, the user could drink successfully using three MI commands in sequence (upper-arm MI to adjacent the robotic arm, hand MI to grasp and forearm MI to twist the robotic arm to assist the user to drink). Theoretically, the shortest control time for one drinking task was 19 seconds.

The exact location of the object was derived by the object detection model YOLO \cite{yolo} from the Kinect V2 RGB-D sensor, as in our previous study \cite{mine_2019SMC}. However, for the case of unexpected decoding results, the user could restore the previous status by performing eye blinking twice; a node horizontally was used as a veto function to initialize the robotic arm, and this counted as a failure of the system.

Table \RomanNumeralCaps{3} presents the success rate of online experiments on drinking water using the MI-based robotic arm and control commands and average control time of each drinking task. Subject 24, who shows the highest performance in the three sessions, could drink nine times successfully, and each of the drinking tasks took 55.73 seconds on an average. Five subjects could achieve a success rate of 78.00\% and each of the tasks contained 9.9 commands on an average.

\section{Discussion}
\subsection{Few-shot Learning Approaches on BCI}

The time-consuming data acquisition process of the BCI system hinders the construction of big data and performance of online tasks. Therefore, many studies attempted to overcome this limitation by using small amounts of various user's data rather than a large number of individuals \cite{Kwon_TNNLS, inde_1, transfer_inde}. However, conventional studies of this well-known subject independent approach are limited in the rough classes such as left and right hand MI because of its low performance. As the purpose of the subject-independent approach is to retrieve common features from various people, the performance is lower than that of the subject dependent approach \cite{Kwon_TNNLS,fahimi2019inter}. Hence, we imported few-shot learning approaches from the vision domain \cite{relationNet, MatchNetwork, PrototypicalNetwork}. Using this approach, we expected to extract discriminant subject dependent features within a few EEG data, while preserving the performance.

\begin{figure*}[!t]
\begin{center}
\includegraphics[width=0.9\textwidth]{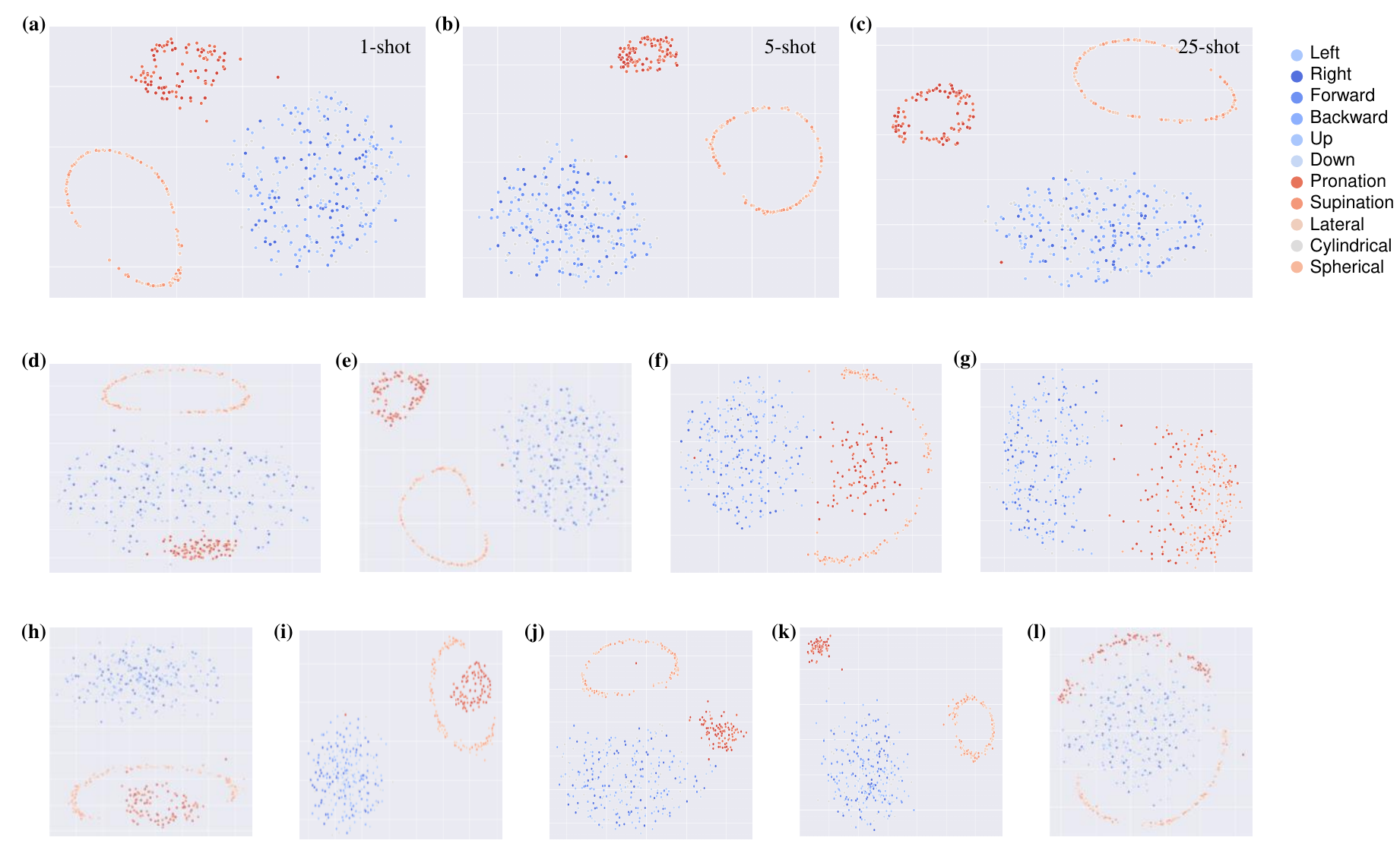}
\caption{Feature space of encoded features for Sub 22 at session 2 using T-SNE. (a)-(c) represents the encoded feature after 1-, 5-, and 25-shot setting, respectively. (d)-(l) depicts the feature space of encoded features of each group of nine trained by 5-shot setting as in (b).}
\label{fig6}
\end{center}
\end{figure*}
As deep learning approaches in the BCI field, such as EEGNet\cite{eegnet} and DeepConvNet\cite{schirrmeister2017deep} contain more parameters that are required to be trained in the case of a machine learning approach such as FBCSP, a deep learning approach suffers in training the model. Therefore, the machine learning approach outperforms the conventional deep learning approach in 5-shot and 25-shot settings. On the contrary, the proposed GRN method could outperform both deep learning and machine learning, irrespective of the training amounts by constructing the correlated feature group and gradually comparing the groups. Furthermore, the metric-based approach of few-shot learning allows the model to choose a comparatively adjacent class related to user intention.

We proposed the first few-shot learning approach in the BCI robotic arm system. A large amount of labeled data in the vision domain allows stable training of the model in deep learning. On the contrary, constructing a large labeled dataset of an individual in the BCI system to perform intuitive online control is nearly impossible as the brain signal of an individual can change owing to the different physiological and psychological characteristics at different times. Therefore, we propose adopting a few-shot learning approach as a further study of the BCI system rather than extracting common features of all the people or performing a time-consuming data acquisition process.

\subsection{Intuitive Commands for BCI-based Robotic Arm Control}

The definition of the term "intuitive" is based on feelings rather than facts or proof; therefore, intuitive MI commands of the BCI system can be varied according to the circumstances that the user faces, and it is impossible to collect all the intuitive MIs in diverse circumstances. Therefore, we used three representative classes that corresponded to each sub-part of the upper extremity and selected eight additional candidate classes that possibly occurred during the online robotic arm control.

The ultimate goal of the intuitive robotic arm is to control the robotic arm without any sense of displacement. The low SNR of the EEG signals and limitless degrees of freedom (DoF) in the upper extremity hinder its ultimate goal. Therefore, in this study, we aimed to ensure the user's intuitive MIs on the multivariate real-time environments rather than limit the user's free will into specific classes by focusing on the sub-part of the upper extremity. For intuitive robotic arm control, asynchronous control is necessary. However, the performance of the proposed GRN method with five training trials was not high enough to realize safe asynchronous robotic arm control. Hence, we decide to perform cue-based robotic arm control with five training trials as a first step towards the realization of asynchronous robotic arm control with small amounts of EEG data.

During the online robotic arm control, we instructed the user to perform intuitive MI command to drink water not only using three representative classes acquired in the data acquisition process but also various intuitive commands. All five subjects could drink water successfully using the robotic arm with an average success rate of 78.00\%. However, the single sub-part movement of the upper extremity limited the DoF of the user. To restore the role of the upper limb successfully, the BCI needs to design a new approach that can decode complicate movements of multiple sub-parts simultaneously, such as hand grasping while arm reaching.

\begin{figure*}[!t]
\begin{center}
\includegraphics[height=0.45\textheight, width=\textwidth]{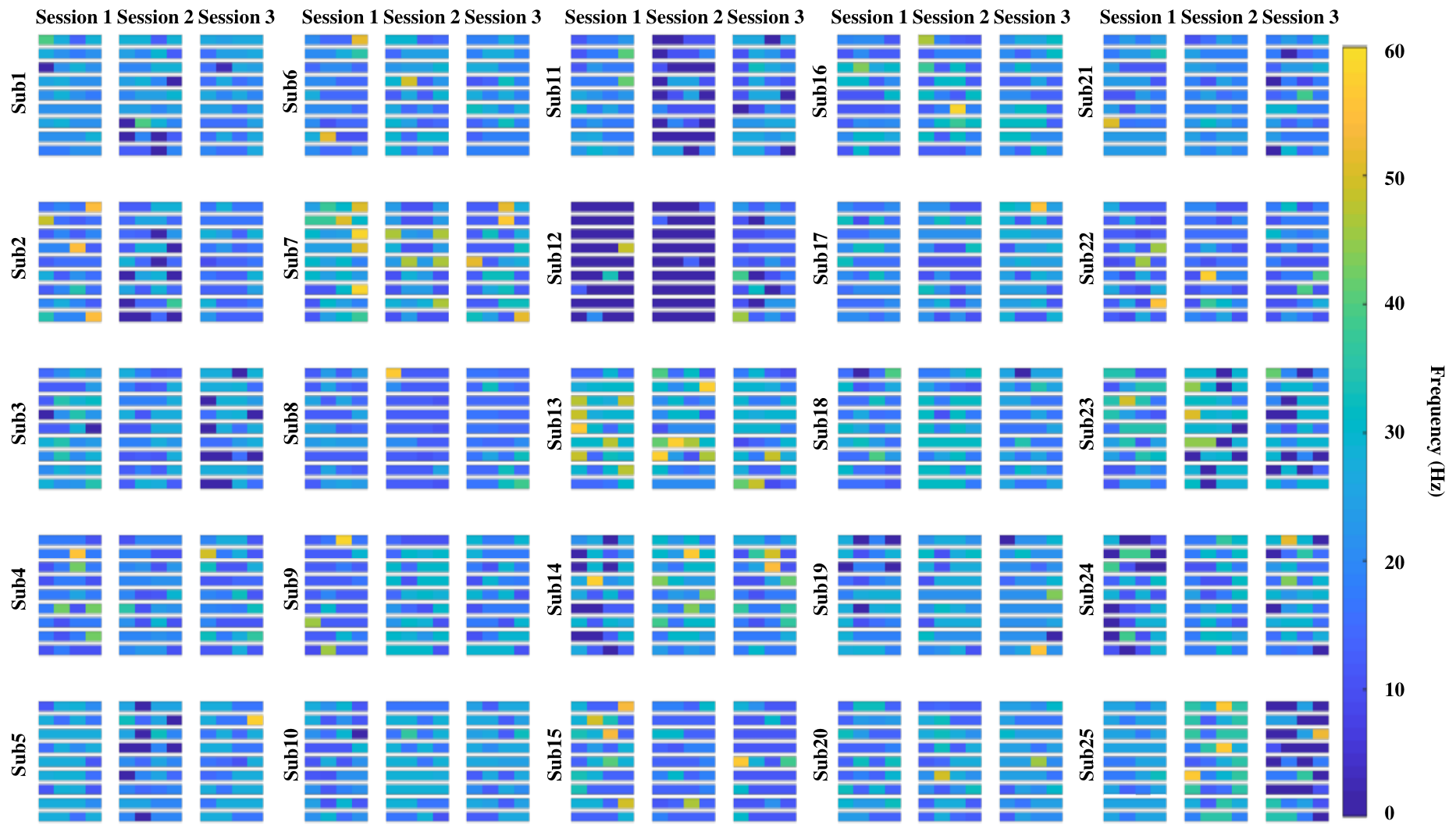}
\caption{Spectral information of each grouped feature after the first channel-wise convolutional layer trained by five training trials. The fast fourier transform is used on the feature map after the channel-wise CNN.}
\label{fig7}
\end{center}
\end{figure*}

\subsection{Comparison of Grouped Information}
The proposed GRN method focused on constructing nine groups containing four temporal or spectral patterns, and gradually comparing two groups from each encoder. By visualizing the encoded feature and its spectral information, we attempted to understand how the model trained with small amounts of EEG data.

Fig. 6. presents the feature space of encoded features from the encoder. The dimensions of the encoded features were reduced using t-stochastic neighbor embedding (T-SNE) to the two-dimensional features space. Both representative and candidate classes were encoded and projected to the feature space. (a), (b), and (c) were the output of the encoder trained by 1, 5, and 25 training trials of Sub 22, respectively. All three models could encode three sub-parts discriminantely on both representative and untrained candidate classes. (d)-(l) represent the encoded features of nine groups of the single encoder. The proposed GRN method attempted to use various feature groups for complementary interaction. As can be inferred from Fig. 6 (d)$\sim$(l), some groups may confuse this owing to non-stationary EEG data, but the relation module of the GRN can compensate for the loss or distorted information using other spectral or temporal groups.

Fig. 7 is the spectral information after the first layer of the encoder. As all the convolutional layers consisted of depthwise CNN and gradually combined at the relation module, the spectral or temporal information is preserved by the group. However, the temporal information of the MI has not been discovered, except for a few patterns such as event-related desynchronization (ERD) or event-related synchronization (ERS) \cite{lotze2006motor, pfurtscheller2001motor}; therefore we focused on visualizing the spectral information, using the fast Fourier transform (FFT). The peak value of the output of FFT is colored as in Fig. 7;approximately 68.59\% of feature components were related to the beta rhythms ([12-30] Hz). Moreover, all the subjects except Sub 12 retrieved the meaningful information from the beta rhythms. The model for Sub 12 extracted the information from the delta band ([0-4] Hz), which may contain MRCP features.

\subsection{Study Limitations and Future Works}
Our study has several limitations that call for future investigations. First, the decoding performance in the 5-shot setting GRN was not sufficient to realize a stable robotic arm control. There are several approaches in few-shot learning such as recurrent neural network (RNN) memory-based approaches\cite{RNN_1,RNN_2Tmp} and fine-tuning approaches\cite{optim_1,optim_2_Tmp}. These use of these approaches may be more appropriate in the BCI domain than in the metric-based GRN. Second, this study is the first attempt at verifying the feasibility of intuitive online robotic arm control with a few EEG data; therefore we performed cue-based online robotic arm control rather than asynchronous control. The continuous control is considered to facilitate the extension of BCI toward the realistic control of physical devices in home and clinical settings \cite{binhe_continuous,leeb2015towards}.
    
\section{Conclusion}
In this study, we proposed the GRN to decode the intuitive MI decoding method for controlling the online robotic arm with a few EEG data. The grouped spectral and temporal feature allows the model to compensate for the low SNR. We verified that the GRN could outperform conventional methods in both representative and untrained candidate classes in proper sub-parts than the conventional methods. 

In conclusion, we constructed the intuitive robotic arm control system with a few training trials using EEG signals. Our study demonstrates superior performance and promising approaches to the few-shot learning method in both offline and online BCI system. This study could pave the way for an intuitive upper extremity MI online robotic arm control within a few EEG data.

\bibliographystyle{IEEEtran}
\bibliography{Manuscript_arxiv}

\begin{thebibliography}{10}
\providecommand{\url}[1]{#1}
\csname url@samestyle\endcsname
\providecommand{\newblock}{\relax}
\providecommand{\bibinfo}[2]{#2}
\providecommand{\BIBentrySTDinterwordspacing}{\spaceskip=0pt\relax}
\providecommand{\BIBentryALTinterwordstretchfactor}{4}
\providecommand{\BIBentryALTinterwordspacing}{\spaceskip=\fontdimen2\font plus
\BIBentryALTinterwordstretchfactor\fontdimen3\font minus
  \fontdimen4\font\relax}
\providecommand{\BIBforeignlanguage}[2]{{%
\expandafter\ifx\csname l@#1\endcsname\relax
\typeout{** WARNING: IEEEtran.bst: No hyphenation pattern has been}%
\typeout{** loaded for the language `#1'. Using the pattern for}%
\typeout{** the default language instead.}%
\else
\language=\csname l@#1\endcsname
\fi
#2}}
\providecommand{\BIBdecl}{\relax}
\BIBdecl

\bibitem{wolpaw2002brain}
J.~R. Wolpaw, N.~Birbaumer, D.~J. McFarland, G.~Pfurtscheller, and T.~M.
  Vaughan, ``Brain--computer interfaces for communication and control,''
  \emph{Clin. Neurophysiol.}, vol. 113, no.~6, pp. 767--791, 2002.

\bibitem{abiri2018comprehensive}
R.~Abiri, S.~Borhani, E.~W. Sellers, Y.~Jiang, and X.~Zhao, ``A comprehensive
  review of {EEG}-based brain-computer interface paradigms,'' \emph{J. Neural.
  Eng.}, vol.~16, no.~1, p. 011001, 2019.

\bibitem{ang2017eeg}
K.~K. Ang and C.~Guan, ``{EEG}-based strategies to detect motor imagery for
  control and rehabilitation,'' \emph{IEEE Trans. Neural Syst. Rehabil. Eng.},
  vol.~25, no.~4, pp. 392--401, 2017.

\bibitem{thirdarm}
C.~I. Penaloza and S.~Nishio, ``B{MI} control of a third arm for
  multitasking,'' \emph{Sci. Robot.}, vol.~3, no.~20, p. eaat1228, 2018.

\bibitem{drone}
N.~Holm \emph{et~al.}, ``An improved five class {MI} based {BCI} scheme for
  drone control using filter bank {CSP},'' in \emph{Int. Winter Conf. on
  Brain-Computer Interface (BCI)}, 2019, pp. 1--6.

\bibitem{lee2018high}
M.-H. Lee, J.~Williamson, D.-O. Won, S.~Fazli, and S.-W. Lee, ``A high
  performance spelling system based on {EEG}-{EOG} signals with visual
  feedback,'' \emph{IEEE Trans. Neural Syst. Rehabil. Eng.}, vol.~26, no.~7,
  pp. 1443--1459, 2018.

\bibitem{BCIReview}
L.~F. Nicolas-Alonso and J.~Gomez-Gil, ``Brain computer interfaces, a review,''
  \emph{Sensors}, vol.~12, no.~2, pp. 1211--1279, 2012.

\bibitem{jeong2020decoding}
J.-H. Jeong, N.-S. Kwak, C.~Guan, and S.-W. Lee, ``Decoding movement-related
  cortical potentials based on subject-dependent and section-wise spectral
  filtering.'' \emph{IEEE Trans. Neural Syst. Rehabil. Eng.}, vol.~28, no.~3,
  pp. 687--698, 2020.

\bibitem{Kwon_TNNLS}
O.-Y. Kwon, M.-H. Lee, C.~Guan, and S.-W. Lee, ``Subject-independent
  brain-computer interfaces based on deep convolutional neural networks,''
  \emph{IEEE Trans. Neural Netw. Learn. Syst.}, pp. 1--14, 2019.

\bibitem{MI_Suk}
H.-I. Suk and S.-W. Lee, ``Subject and class specific frequency bands selection
  for multiclass motor imagery classification,'' \emph{Int. J. Imaging Syst.
  Technol.}, vol.~21, no.~2, pp. 123--130, 2011.

\bibitem{li2018hybrid}
J.~Li, Z.~L. Yu, Z.~Gu, W.~Wu, Y.~Li, and L.~Jin, ``A hybrid network for {ERP}
  detection and analysis based on restricted {B}oltzmann machine,'' \emph{IEEE
  Trans. Neural Syst. Rehabil. Eng.}, vol.~26, no.~3, pp. 563--572, 2018.

\bibitem{lotze2006motor}
M.~Lotze and U.~Halsband, ``Motor imagery,'' \emph{J. Physiol. Paris}, vol.~99,
  no. 4-6, pp. 386--395, 2006.

\bibitem{jafarifarmand2017new}
A.~Jafarifarmand, M.~A. Badamchizadeh, S.~Khanmohammadi, M.~A. Nazari, and
  B.~M. Tazehkand, ``A new self-regulated neuro-fuzzy framework for
  classification of {EEG} signals in motor imagery {BCI},'' \emph{IEEE Trans.
  Fuzzy Syst.}, vol.~26, no.~3, pp. 1485--1497, 2017.

\bibitem{kwak2019error}
N.-S. Kwak and S.-W. Lee, ``Error correction regression framework for enhancing
  the decoding accuracies of ear-{EEG} brain-computer interfaces,'' \emph{IEEE
  Trans. on Cybernetics}, 2019.

\bibitem{leeb2015towards}
R.~Leeb \emph{et~al.}, ``Towards independence: {a} {BCI} telepresence robot for
  people with severe motor disabilities,'' \emph{Proc. IEEE}, vol. 103, no.~6,
  pp. 969--982, 2015.

\bibitem{mine_2019SMC}
K.-H. Shim, J.-H. Jeong, B.-H. Kwon, B.-H. Lee, and S.-W. Lee, ``Assistive
  robotic arm control based on brain-machine interface with vision guidance
  using convolution neural network,'' in \emph{IEEE Int. Conf. Syst., Man, and
  Cybern. (SMC)}, Oct 2019, pp. 2785--2790.

\bibitem{schwarz2017decoding}
A.~Schwarz, P.~Ofner, J.~Pereira, A.~I. Sburlea, and G.~R. M{\"u}ller-Putz,
  ``Decoding natural reach-and-grasp actions from human {EEG},'' \emph{J.
  Neural Eng.}, vol.~15, no.~1, p. 016005, 2017.

\bibitem{jeong2020Brain}
J.-H. Jeong, K.-H. Shim, D.-J. Kim, and S.-W. Lee, ``Brain-controlled robotic
  arm system based on multi-directional {CNN}-{B}i{LSTM} network using {EEG}
  signals.'' \emph{IEEE Trans. Neural Syst. Rehabil. Eng.}, 2020.

\bibitem{xu2019shared}
Y.~Xu \emph{et~al.}, ``Shared control of a robotic arm using non-invasive
  brain-computer interface and computer vision guidance,'' \emph{Robot. Auton.
  Syst.}, vol. 115, pp. 121--129, 2019.

\bibitem{shared}
S.~Schr{\"o}er \emph{et~al.}, ``An autonomous robotic assistant for drinking,''
  in \emph{Proc. Int. Conf. Robot. Auton. (ICRA)}, 2015, pp. 6482--6487.

\bibitem{reduceTrial}
Y.~Jiao \emph{et~al.}, ``Sparse group representation model for motor imagery
  {EEG} classification,'' \emph{IEEE J. Biomed. Health Inform.}, vol.~23,
  no.~2, pp. 631--641, 2018.

\bibitem{craik2019deep}
A.~Craik, Y.~He, and J.~L.~P. Contreras-Vidal, ``Deep learning for
  electroencephalogram ({EEG}) classification tasks: A review,'' \emph{J.
  Neural Eng.}, vol.~16, no.~3, p. 031001, 2019.

\bibitem{singh2019reduce}
A.~Singh, S.~Lal, and H.~W. Guesgen, ``Reduce calibration time in motor imagery
  using spatially regularized symmetric positives-definite matrices based
  classification,'' \emph{Sensors}, vol.~19, no.~2, p. 379, 2019.

\bibitem{eegnet}
V.~J. Lawhern, A.~J. Solon, N.~R. Waytowich, S.~M. Gordon, and C.~P. Hung,
  ``{EEGN}et: {A} compact convolutional neural network for {EEG}-based
  brain-computer interfaces,'' \emph{J. Neural Eng.}, vol.~15, no.~5, p.
  056013, 2018.

\bibitem{schirrmeister2017deep}
R.~T. Schirrmeister \emph{et~al.}, ``Deep learning with convolutional neural
  networks for {EEG} decoding and visualization,'' \emph{Hum. Brain Mapp.},
  vol.~38, no.~11, pp. 5391--5420, 2017.

\bibitem{tayeb2019validating}
Z.~Tayeb \emph{et~al.}, ``Validating deep neural networks for online decoding
  of motor imagery movements from {EEG} signals,'' \emph{Sensors}, vol.~19,
  no.~1, p. 210, 2019.

\bibitem{relationNet}
F.~Sung, Y.~Yang, L.~Zhang, T.~Xiang, P.~H. Torr, and T.~M. Hospedales,
  ``Learning to compare: {R}elation network for few-shot learning,'' in
  \emph{Proc. IEEE Conf. Comput. Vis. Pattern Recognit. (CVPR)}, June 2018.

\bibitem{PrototypicalNetwork}
J.~Shell, K.~Swersky, and R.~Zemel, ``Prototypical networks for few-shot
  learning,'' in \emph{Proc. Adv. Neural Inf. Process. Syst. (NIPS)}, 2017, pp.
  4077--4087.

\bibitem{Hong_Few}
H.-G. Jung and S.-W. Lee, ``Few-shot learning with geometric constraints,''
  \emph{IEEE Trans. Neural Netw. Learn. Syst.}, 2020.

\bibitem{jeong2019classification}
J.-H. Jeong, B.-W. Yu, D.-H. Lee, and S.-W. Lee, ``Classification of drowsiness
  levels based on a deep spatio-temporal convolutional bidirectional {LSTM}
  network using electroencephalography signals,'' \emph{Brain Sci.}, vol.~9,
  no.~12, p. 348, 2019.

\bibitem{jiao2018deep}
Z.~Jiao, X.~Gao, Y.~Wang, J.~Li, and H.~Xu, ``Deep convolutional neural
  networks for mental load classification based on {EEG} data,'' \emph{Pattern
  Recognit.}, vol.~76, pp. 582--595, 2018.

\bibitem{zhang2019learning}
P.~Zhang, X.~Wang, W.~Zhang, and J.~Chen, ``Learning spatial-spectral-temporal
  {EEG} features with recurrent 3{D} convolutional neural networks for
  cross-task mental workload assessment,'' \emph{IEEE Trans. Neural Syst.
  Rehabil. Eng.}, vol.~27, no.~1, pp. 31--42, 2019.

\bibitem{lu2017deep}
N.~Lu, T.~Li, X.~Ren, and H.~Miao, ``A deep learning scheme for motor imagery
  classification based on restricted {B}oltzmann machines,'' \emph{IEEE Trans.
  Neural Syst. Rehabil. Eng.}, vol.~25, no.~6, pp. 566--576, 2017.

\bibitem{zhang2019novel}
Z.~Zhang \emph{et~al.}, ``A novel deep learning approach with data augmentation
  to classify motor imagery signals,'' \emph{IEEE Access}, vol.~7, pp.
  15\,945--15\,954, 2019.

\bibitem{kaya2018large}
M.~Kaya \emph{et~al.}, ``A large electroencephalographic motor imagery dataset
  for electroencephalographic brain computer interfaces,'' \emph{Sci. Data},
  vol.~5, p. 180211, 2018.

\bibitem{lee2019eeg}
M.-H. Lee \emph{et~al.}, ``{EEG} dataset and openbmi toolbox for three {BCI}
  paradigms: an investigation into {BCI} illiteracy,'' \emph{GigaScience},
  vol.~8, no.~5, p. giz002, 2019.

\bibitem{MatchNetwork}
O.~Vinyals, C.~Blundell, T.~Lillicrap, K.~Kavukcuoglu, and D.~Wiestra,
  ``Matching networks for one shot learning,'' in \emph{Proc. Adv. Neural Inf.
  Process. Syst. (NIPS)}, 2016, pp. 3630--3638.

\bibitem{koch2015siamese}
G.~Koch, R.~Zemel, and R.~Salakhutdinov, ``Siamese neural networks for one-shot
  image recognition,'' in \emph{Proc. Int. Conf. Mach. Learn. (ICML)}, vol.~2,
  2015.

\bibitem{NNLS_CGuan}
S.~Sakhavi, C.~Guan, and S.~Yan, ``Learning temporal information for
  brain-computer interface using convolutional neural networks,'' \emph{IEEE
  Trans. Neural Netw. Learn. Syst.}, vol.~29, no.~11, pp. 5619--5629, Nov 2018.

\bibitem{PFURTSCHELLER1997642}
G.~Pfurtscheller, C.~Neuper, D.~Flotzinger, and M.~Pregenzerb, ``E{EG}-based
  discrimination between imagination of right and left hand movement,''
  \emph{Electroencephalogr. Clin. Neurophysiol.}, vol. 103, no.~6, pp. 642 --
  651, 1997.

\bibitem{Blankertz_NeuroImage}
B.~Blankertz, G.~Dornhege, M.~Krauledat, K.-R. M{\"u}ller, and G.~Curio, ``The
  non-invasive berlin brain–computer interface: {F}ast acquisition of
  effective performance in untrained subjects,'' \emph{Neuroimage}, vol.~37,
  no.~2, pp. 539 -- 550, 2007.

\bibitem{LotteReview}
F.~Lotte, M.~Congedo, A.~Le\textsc{\char13}cuyer, F.~Lamarche, and B.~Arnaldi,
  ``A review of classification algorithms for {EEG}-based
  brain{\textendash}computer interfaces,'' \emph{J. Neural Eng.}, vol.~4,
  no.~2, pp. R1--R13, jan 2007.

\bibitem{kingma2014adam}
D.~P. Kingma and J.~Ba, ``Adam: {A} method for stochastic optimization,''
  \emph{arXiv preprint arXiv:1412.6980}, 2014.

\bibitem{review_2019}
Y.~Roy, H.~Banville, I.~Albuquerque, A.~Gramfort, T.~H. Falk, and J.~Faubert,
  ``Deep learning-based electroencephalography analysis: {A} systematic
  review,'' \emph{J. Neural Eng.}, 2019.

\bibitem{cole2019cycle}
S.~R. Cole and B.~Voytek, ``Cycle-by-cycle analysis of neural oscillations,''
  \emph{J. Neurophysiol.}, vol. 122, no.~2, pp. 849--861, 2019.

\bibitem{gramfort2013time}
A.~Gramfort, D.~Strohmeier, J.~Haueisen, M.~H{\"a}m{\"a}l{\"a}inen, and
  M.~Kowalski, ``Time-frequency mixed-norm estimates: {S}parse {M}/{EEG}
  imaging with non-stationary source activations,'' \emph{Neuroimage}, vol.~70,
  pp. 410--422, 2013.

\bibitem{fbcsp}
K.~K. Ang, Z.~Y. Chin, H.~Zhang, and C.~Guan, ``Filter bank common spatial
  pattern ({FBCSP}) in brain-computer interface,'' in \emph{IEEE Int. Joint
  Conf. on Neural Netw.}, 2008, pp. 2390--2397.

\bibitem{yolo}
J.~Redmon, S.~Divvala, R.~Girshick, and A.~Farhadi, ``You only look once:
  {U}nified, real-time object detection,'' in \emph{Proc. IEEE Conf. Comput.
  Vis. Pattern Recognit. (CVPR)}, 2016, pp. 779--788.

\bibitem{inde_1}
A.~M. Ray \emph{et~al.}, ``A subject-independent pattern-based brain-computer
  interface,'' \emph{Front. Behav. Neurosci.}, vol.~9, p. 269, 2015.

\bibitem{transfer_inde}
V.~Jayaram, M.~Alamgir, Y.~Altun, B.~Scholkopf, and M.~Grosse-Wentrup,
  ``Transfer learning in brain-computer interfaces,'' \emph{IEEE Comput.
  Intell. M.}, vol.~11, no.~1, pp. 20--31, 2016.

\bibitem{fahimi2019inter}
F.~Fahimi, Z.~Zhang, W.~B. Goh, T.-S. Lee, K.~K. Ang, and C.~Guan,
  ``Inter-subject transfer learning with an end-to-end deep convolutional
  neural network for {EEG}-based {BCI},'' \emph{J. Neural Eng.}, vol.~16,
  no.~2, p. 026007, 2019.

\bibitem{pfurtscheller2001motor}
G.~Pfurtscheller and C.~Neuper, ``Motor imagery and direct brain-computer
  communication,'' \emph{Proc. IEEE}, vol.~89, no.~7, pp. 1123--1134, 2001.

\bibitem{RNN_1}
T.~Munkhdalai and H.~Yu, ``Meta networks,'' in \emph{Proc. Int. Conf. Mach.
  Learn. (ICML)}, 2017, pp. 2554--2563.

\bibitem{RNN_2Tmp}
A.~Santoro, S.~Bartunov, M.~Botvinick, D.~Wierstra, and T.~Lillicrap,
  ``Meta-learning with memory-augmented neural networks,'' in \emph{Proc. Int.
  Conf. Mach. Learn. (ICML)}, 2016, pp. 1842--1850.

\bibitem{optim_1}
C.~Finn, P.~Abbeel, and S.~Levine, ``Model-agnostic meta-learning for fast
  adaptation of deep networks,'' in \emph{Proc. Int. Conf. Mach. Learn.
  (ICML)}, 2017, pp. 1126--1135.

\bibitem{optim_2_Tmp}
S.~Ravi and H.~Larochelle, ``Optimization as a model for few-shot learning,''
  in \emph{Proc. Int. Conf. Learn. Represent. (ICLR)}, 2017.

\bibitem{binhe_continuous}
B.~J. Edelman \emph{et~al.}, ``Noninvasive neuroimaging enhances continuous
  neural tracking for robotic device control,'' \emph{Sci. Robot.}, vol.~4,
  no.~31, p. eaaw6844, 2019.

\end{thebibliography}

\end{document}